\documentclass{article}

\usepackage{microtype}
\usepackage{graphicx}
\usepackage{subfigure}
\usepackage{booktabs} % for professional tables
\usepackage{multirow}

\usepackage{hyperref}

\usepackage[accepted]{icml2024}

\usepackage{amsmath}
\usepackage{amssymb}
\usepackage{mathtools}
\usepackage{amsthm}

\usepackage[capitalize,noabbrev]{cleveref}

\theoremstyle{plain}

\theoremstyle{definition}

\theoremstyle{remark}

\usepackage[textsize=tiny]{todonotes}

\icmltitlerunning{Local CorEx}

\begin{document}

\twocolumn[
\icmltitle{Exploring higher-order neural network node interactions with total correlation}




\begin{icmlauthorlist}
\icmlauthor{Thomas Kerby}{yyy}
\icmlauthor{Teresa White}{yyy}
\icmlauthor{Kevin Moon}{yyy}
\end{icmlauthorlist}

\icmlaffiliation{yyy}{Department of Mathematics and Statistics, Utah State University, Logan, UT 84322, United States}

\icmlcorrespondingauthor{Kevin Moon}{kevin.moon@usu.edu}

\icmlkeywords{Variable Interaction Detection, Interpretable Machine Learning, Higher Order Interaction}

\vskip 0.3in
]



\printAffiliationsAndNotice{}  % leave blank if no need to mention equal contribution

\begin{abstract}
In domains such as ecological systems, collaborations, and the human brain the variables interact in complex ways. Yet accurately characterizing higher-order variable interactions (HOIs) is a difficult problem that is further exacerbated when the HOIs change across the data. To solve this problem we propose a new method called Local Correlation Explanation (CorEx) to capture HOIs at a local scale by first clustering data points based on their proximity on the data manifold. We then use a multivariate version of the mutual information called the total correlation, to construct a latent factor representation of the data within each cluster to learn the local HOIs. We use Local CorEx to explore HOIs in synthetic and real world data to extract hidden insights about the data structure. Lastly, we demonstrate Local CorEx's suitability to explore and interpret the inner workings of trained neural networks.
\end{abstract}

\section{Introduction}
\label{introduction}

Explainability and interpretability are fundamental for trust and adoption of machine learning algorithms across various applications. Yet understanding the decision process of even relatively simple neural networks on complex data can be difficult to achieve. A recent survey on inner neural network interpretability techniques by \cite{rauker_2023} divides interpretability tools into four main groups based on which part of the network's computational graph they explain: weights, neurons, subnetworks, and latent representations. Each of these groups can be categorized further as intrinsic and post hoc, where the former involves training models in a manner to be more easily interpretable, and the latter aims to interpret the model after training is completed. Additional work focused on discovering novel behaviors in neural networks include identifying where knowledge is stored in LLM weights \cite{dai2021knowledge,meng2022locating}, labeling nodes with natural language descriptions \cite{hernandez2022natural}, and editing classifiers to interpret a concept in a new manner given an example \cite{santurkar_2021}. All of these methods require either an input of interest or the class labels and are thus supervised.

In response to the challenges posed by understanding neural networks and analyzing higher-order variable interactions (HOIs), we present Local CorEx, a novel post hoc method suitable for exploring model weights, nodes, subnetworks, and latent representations in an unsupervised manner. Here we focus our attention on analyzing groups of hidden nodes and latent representations. To the best of our knowledge, our work marks the first post hoc method to do so in an unsupervised manner and includes the option to easily incorporate label information. Additionally, our approach extends to analyzing HOIs within the data.

Higher-order models capture all possible interactions up to order $k$ among the variables in a study. The interactions can be considered either with or without the context of a response or label. In the latter case, we care to understand the nature of the data and how the measured features relate to one another, such as the inner workings of deep neural networks. Variable interactions have long been studied using tools such as the Pearson correlation \cite{pearson_correlation}, the Spearman correlation \cite{spearman_correlation}, mutual information \cite{mutual_information}, and total correlation (multivariate mutual information) \cite{estimating_total_correlation_MI}. Such tools have proven useful in exploring and understanding datasets. Unfortunately, using said tools on datasets with large numbers of variables does not scale well as the number of possible HOIs grows as $O(2^n)$ where $n$ is the number of variables. Because of this and the difficulty in interpreting coefficients associated with HOIs \cite{Neter1983AppliedLR} advise caution for including HOIs in regression models.

Despite the difficulty of learning HOIs, it is becoming increasingly apparent that they play a key role in some complex systems \cite{zhang2023higher,boccaletti2023structure}. Domains such as ecological systems \cite{grilli2017higher, levine_2022}, collaborations \cite{civilini2023explosive}, and the human brain \cite{petri2014homological, giusti2016two} have all been shown to be impacted by HOIs. 

One approach to learning HOIs has focused on transforming complex and high-dimensional data into a simpler representation \cite{steeg2015maximally}. Learning such representations of the data can help to extract useful information when building classifiers or predictors \cite{bengio2014representation}. The principle of Cor-relation Ex-planation (CorEx) was introduced to construct informative representations that provide valuable information about relationships between variables in high-dimensional data \cite{steeg2017discov_struc}. A particular variant, called Linear CorEx, estimates multivariate Gaussian distributions by identifying independent latent factors that explain correlations among observed variables \cite{linear_corex}. It incorporates a modular inductive prior, favoring models where the covariance matrix is block-diagonal, indicating clusters governed by a few latent factors. Another variant is called Bio CorEx, which focuses primarily on handling challenges inherent in several biomedical problems: missing data, continuous variables, and severely under-sampled data \cite{Pepke2017}.

While both Linear and Bio CorEx have been used successfully, they fail to consider variable interactions may vary across the data manifold. For example, in a classification problem,  interactions are often not shared across classes because the data lie on separate data manifolds. Thus Linear and Bio CorEx may only find the interactions that span classes while obscuring or distorting class-specific interactions. Local CorEx solves this problem by partitioning the data prior to estimating variable interactions.  

Our contributions are: 1) we derive a novel method for estimating local variable interactions (Section \ref{Local Corex}); 2) we show on synthetic data that Local CorEx is robust to hyperparameter selection and outperforms previous works when HOIs vary across the data manifold (Section \ref{sub:ablation}); 3) we demonstrate Local CorEx's utility in extracting plausible HOIs from real-world data (Sections \ref{communities_dataset} \& \ref{sub:MNIST}); 4) we interpret the inner workings of a neural network classifier using Local CorEx and discover sets of hidden nodes needed to accurately classify a local cluster while leaving accuracies for other clusters relatively unaffected (Section \ref{sub:neural_net}).

\section{Local CorEx} \label{Local Corex}

\begin{figure*}
    \centering
    \includegraphics[width=0.9\textwidth]{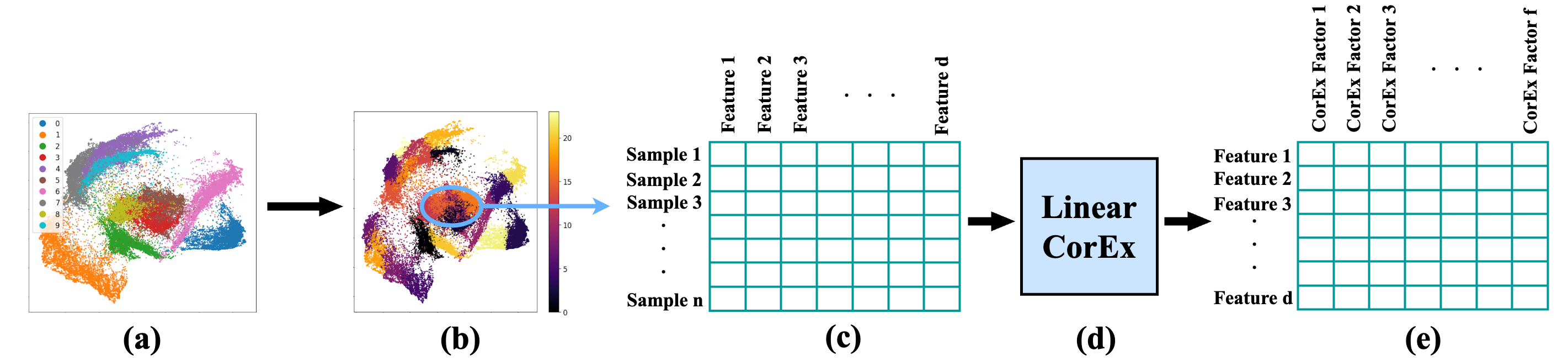}
    \caption{Overview of the Local CorEx algorithm. \textit{(a)} PHATE visualization of the MNIST dataset. \textit{(b)} $k$-means clustering is applied to the PHATE embedding to generate the local clusters. \textit{(c-d)} A cluster is chosen and passed through Linear CorEx. \textit{(e)} We visualize the mutual information between the learned CorEx latent factors and the original features to identify HOIs.}
    \label{fig:main_local_corex_fig}
\end{figure*}

Figure~\ref{fig:main_local_corex_fig} gives an overview of Local CorEx. Local CorEx first uses PHATE~\cite{Moon2019} to create a low-dimensional embedding of the data that preserves the local and global structure of the data manifold(s). PHATE is a dimensionality reduction technique that is specialized for data visualization \cite{Moon2019}. It learns the local structure via a specialized kernel function. The pairwise kernel matrix is row-normalized to create a Markov transition matrix, called the diffusion operator \cite{coifman2006diffusion}.  The global structure of the data is learned via diffusion by simulating $t$-step random walks between data points according to the normalized affinities.  Finally, information distances are calculated between the diffused probabilities, and multidimensional scaling (MDS) \cite{Kruskal1964} is used to preserve these distances in low dimensions. 

To partition the data, we apply $k$-means clustering \cite{kmeans_clust} to the PHATE embedding with where $k$ is chosen to be ``large enough" to ensure heterogeneity among the partitions according to a manual inspection of a 2D PHATE plot. This is akin to spectral clustering \cite{spectral_2019}, which essentially applies $k$-means to the Laplacian Eigenmaps embedding. We choose this clustering approach over other methods because of PHATE's impressive visualization capabilities. After performing the clustering we can easily visualize the data using PHATE and color by clusters to see how the clusters relate to each other and if the partition makes visual sense.

 After the partition has been created, we next apply either Linear or Bio CorEx to each cluster to obtain the local variable interactions. We now describe how Linear and Bio CorEx work. Let $X \equiv X_{1:p} \equiv \{X_1, X_2, \cdots, X_p\}$ denote a vector of $p$ observed variables and let $Z \equiv Z_{1:m} \equiv \{Z_1, Z_2, \cdots, Z_m\}$ denote a vector of $m$ latent variables. Instances of $X$ and $Z$ are denoted in lowercase with $x = (x_1, x_2, \cdots, x_p)$ and $z = (z_1, z_2, \cdots, z_m)$, respectively. We consider several information theoretic measures including differential entropy: $H(X) = -\mathbb{E}[\log{p(x)}]$ ($p(x)$ is the probability density of $X$), mutual information: $I(X;Y) = H(X) + H(Y) - H(X,Y)$, Total Correlation: $TC(X) = \sum_{i=1}^p H(X_i) - H(X)$, and their conditional variants such as $H(X|Z) = \mathbb{E}_z[H(X|Z = z)]$ and $TC(X|Z) = \mathbb{E}_z[TC(X|Z = z)]$. In particular, the total correlation measures the redundancy or dependency among a set of $n$ random variables.

%We use both Linear and Bio CorEx to estimate variable interactions within the partitioned space. A sketch of the main framework for motivating Linear CorEx is provided below with a full derivation provided in \cite{linear_corex}. 
Linear CorEx~\cite{linear_corex} estimates the latent factors $z$ by optimizing the following expression:
\begin{equation}
    \min_{W} TC(X|Z) + TC(Z) + \sum_{i=1}^pQ_i,
\end{equation}
where $z = Wx + \epsilon$, $W \in \mathbf{R}^{m\times p}$, $\epsilon \sim \mathcal{N}(0, \Sigma)$, $\Sigma$ is a diagonal matrix, and the $Q_i$'s are non-negative regularization terms that encourage modular solutions (i.e. solutions with small values of $TC(Z|X_i)$) and only equal 0 when the solution is modular. In essence, Linear CorEx attempts to identify latent factors that constitute a linear combination of the inputs, explain the total correlation in the data, remain independent of one another, and are modular.

%old version from above: In essence, Linear CorEx is trying to find latent factors that are a linear combination of the inputs, explain the total correlation in the data, are independent of one another, and are modular.

In contrast, Bio CorEx creates discrete latent factors with a predetermined finite number of possible states~\cite{Pepke2017} by maximizing the following wrt $z$:
\begin{align}
    TC(X;Z) &\equiv \sum_{i = 1}^nI(X_i:Z) - I(X:Z)\\
    & = TC(X) - TC(X|Z) 
\end{align}
Bio CorEx formulates a tractable lower bound for $TC(X;Z)$ such that it optimizes a single latent factor conditioned on previously found latent factors. It then iterates through the latent factors until it has learned the predetermined number of latent factors. This can be thought of as finding a latent factor that is maximally informative about the data given the previously learned latent factors. 

Both Linear and Bio CorEx are far faster then a brute force search. Bio CorEx has a computational complexity of $O(m*n*N)$ where $m$ is the number of latent factors, $n$ is the number of variables, and $N$ is the number of samples. Although Linear CorEx shares the same computational complexity, its factors can be learned in parallel.

 The total correlation explained by a latent factor is a measure of its importance. We can identify the predicted HOIs by examining the mutual information between each latent factor and the original set of features. Here the magnitude of the mutual information can be used as a proxy for the strength or importance of the feature to the predicted HOI. 

Local CorEx hyperparameters include: the number of dimensions to include in the PHATE embedding of the data, the number of clusters for $k$-means, the number of CorEx latent factors to learn, and for each CorEx factor, a threshold for when a feature is a part of a learned HOI. Hyperparameter selection guidelines are given in Appendix~\ref{sec:hyperparameters}. 

%For a more in-depth analysis of this method see \cite{steeg2015maximally}.

\section{Results} \label{Results}

    \subsection{Ablation Study - Synthetic Data} \label{sub:ablation}

        \begin{table*}[t]
            \caption{Subset of the ablation study results for the sample size of 1000. For all sample sizes see Appendix \ref{appendix_ablation_study}. Based on the metrics, Local CorEx outperforms the global methods when the data have mixed variable interactions. For sufficiently large sample sizes, Local CorEx also outperforms or approximately matches the global methods when interactions are not mixed.}
            \label{tab:ablation_study}
            \begin{center}
            \begin{small}
            \begin{sc}
            \begin{tabular}{llrrrrrrrr}
            \toprule
             & & \multicolumn{4}{c}{Group} & \multicolumn{4}{c}{Top K Latent Factor}\\
            & & \multicolumn{2}{c}{Cosine Dist} & \multicolumn{2}{c}{AUCPRC} & \multicolumn{2}{c}{Cosine Dist} & \multicolumn{2}{c}{AUCPRC}\\
            Data & Method & $\alpha=0.0$ & $\alpha=1.0$ & $\alpha=0.0$ & $\alpha=1.0$ & $\alpha=0.0$ & $\alpha=1.0$ & $\alpha=0.0$ & $\alpha=1.0$ \\
            \midrule
             Disjoint & Linear & 0.516 & 0.408 & 0.606 & 0.604 & 0.412 & 0.401 & 0.701 & 0.640 \\
             & Bio & \textbf{0.414} & 0.474 & \textbf{0.652} & 0.507 & \textbf{0.323} & 0.436 & \textbf{0.758} & 0.617 \\
             & Local Linear & 0.504 & \textbf{0.146} & 0.607 & \textbf{0.902} & 0.420 & 0.031 & 0.691 & 0.999 \\
             & Local Bio & 0.451 & 0.163 & 0.610 & 0.883 & 0.374 & \textbf{0.028} & 0.708 & \textbf{1.000} \\
             \cline{1-10}\\
             Non- & Linear   & \textbf{0.191} & 0.182 & \textbf{0.886} & 0.856 & 0.091 & 0.086 & \textbf{0.992} & 0.959 \\
             Disjoint  & Bio          & 0.203 & 0.276 & 0.869 & 0.773 & \textbf{0.084} & 0.199 & 0.991 & 0.892 \\
             &  Local Linear & 0.197 & \textbf{0.141} & 0.870 & \textbf{0.908} & 0.100 & 0.051 & 0.959 & \textbf{1.000} \\
             & Local Bio    & 0.222 & 0.173 & 0.850 & 0.873 & 0.107 & \textbf{0.039} & 0.965 & \textbf{1.000} \\
            \bottomrule
            \end{tabular}
            \end{sc}
            \end{small}
            \end{center}
        \end{table*}
        
        %To demonstrate the robustness of Local CorEx to chosen hyperparameters, 
        To demonstrate the effectiveness of Local CorEx to learn HOIs we constructed a synthetic dataset containing two clusters where the variable interactions are known and consist of grouped pairwise interactions, since to the best of our knowledge no standard dataset with known HOIs exists. We also compare to standard Linear and Bio CorEx. In this ablation study, we show how all methods perform when we vary the number of chosen latent factors, the difference in interactions between partitions, and the number of data samples. For each simulation setup, we ran 16 replicates. 

        The synthetic dataset is composed of two multivariate normal distributions following two setups. In each setup, we varied the class means between the two distributions using the parameter $\alpha$ to control the degree of separation ($\alpha=0$ when clusters share the same mean and $\alpha=1$ when cluster means are maximally separated). This controlled how close the two clusters are in proximity and as a result how pure the clusters are after partitioning. In the first setup, which we call non-disjoint, the two clusters share the majority of their interactions. In the second setup, which we label as disjoint, none of the interactions match. The interactions are identified by extracting each row from the covariance matrix and then examining all of the nonzero values in the row. To find the set of interactions present in the data we collect all of the unique rows and store them in a set. See Appendix \ref{appendix_ablation_study} for additional details about synthetic data generation.

        To find each predicted HOI we first ran either Linear CorEx or Bio CorEx on all, or a portion, of the simulation data points as shown in Figure \ref{fig:main_local_corex_fig}. Then, for each CorEx latent factor, we examined the mutual information between the learned latent factor and the original covariates in the data. To score how well the predicted HOI matches the ground truth interactions, we used two metrics:  1) the cosine distance and 2) the area under the precision-recall curve. For each of these metrics, we compute two versions. The group version can be viewed from a recall perspective where we average the metrics across all of the interactions with the most similar predicted HOIs (See Algorithm \ref{alg:group_scoring}). The second version we refer to as ``top $k$" is akin to precision. If there are $g$ HOIs and $l$ learned CorEx factors then $k = g$ if $l > g$. Otherwise $k = l$. In this case, we look at the top $k$ CorEx factors and compare them with the closest true HOIs (See Algorithm \ref{alg:top_k_scoring}). The scores for a sample size of $1000$ per cluster are shown in Table \ref{tab:ablation_study}. The results for other sample sizes are in Appendix \ref{appendix_ablation_study}.

        % One takeaway from Table \ref{tab:ablation_study} is that when $\alpha$ is high (partitions are pure), Local CorEx beats its global counterparts even when the majority of HOIs overlap between data clusters (i.e. the non-disjoint setting). When the partitions are not pure ($\alpha=0$) and cluster means are the same, Local CorEx only performs marginally worse mostly due to a decrease in sample size. However, when the interactions don't overlap much between groups there is a lot to gain from using Local CorEx. This suggests that applying Local CorEx is a low-risk high reward trade-off so long as the sample size is sufficiently large for CorEx to learn appropriate latent factors. Appendix \ref{appendix_ablation_study} shows these results for different sample sizes. From it, we see that for the disjoint data, so long as there if sufficient separation between clusters, Local CorEx (regardless of explored sample sizes) outperforms global methods regardless of the sample sizes.

        % Additionally, using Local Linear CorEx or Local Bio CorEx results in fairly similar outcomes despite Bio CorEx being the more flexible method. Given that Bio CorEx does not computationally scale as well as Linear CorEx as the number of samples and features increases (see Figure \ref{fig:timing}) we use Local Linear CorEx for the remainder of the paper and refer to it as Local CorEx. 

        From Table \ref{tab:ablation_study} and its fuller counterpart in Appendix \ref{appendix_ablation_study} we note that global methods, like Linear CorEx and Bio CorEx, suffer when data with mixed variable interactions are added. Partitioning the data prior to running these algorithms can help substantially when ``good" partitions are chosen. Additionally, partitioning when the interactions are mostly the same has little impact on performance except when the sample size is sufficiently small, as we are effectively decreasing the already small sample size. This suggests that applying Local CorEx is a low-risk high reward trade-off so long as the sample size is sufficiently large for CorEx to learn appropriate latent factors. In general, Linear CorEx matches or outperforms Bio CorEx despite Bio CorEx being a more flexible method. This may be because we are using multivariate normal distributions, but it is convenient given that Linear CorEx runs much faster than Bio CorEx as shown in Figure \ref{fig:timing}. For the remainder of the paper, we use a local partitioning of the data and apply Linear CorEx.
        
    \subsection{Communities Dataset} \label{communities_dataset}

        To demonstrate Local CorEx on real data, we used the communities and crime dataset \cite{misc_communities_and_crime_183} from the UC Irvine Machine Learning Repository, which consists of aggregated community data for 1,994 different communities spread across the United States. The data features consist of socio-economic data extracted from the 1990 US Census, law enforcement data extracted from the 1990 US LEMAS survey, and crime data extracted from the 1995 FBI UCR.

\begin{figure}
            \centering
            \includegraphics[width=.47\textwidth]{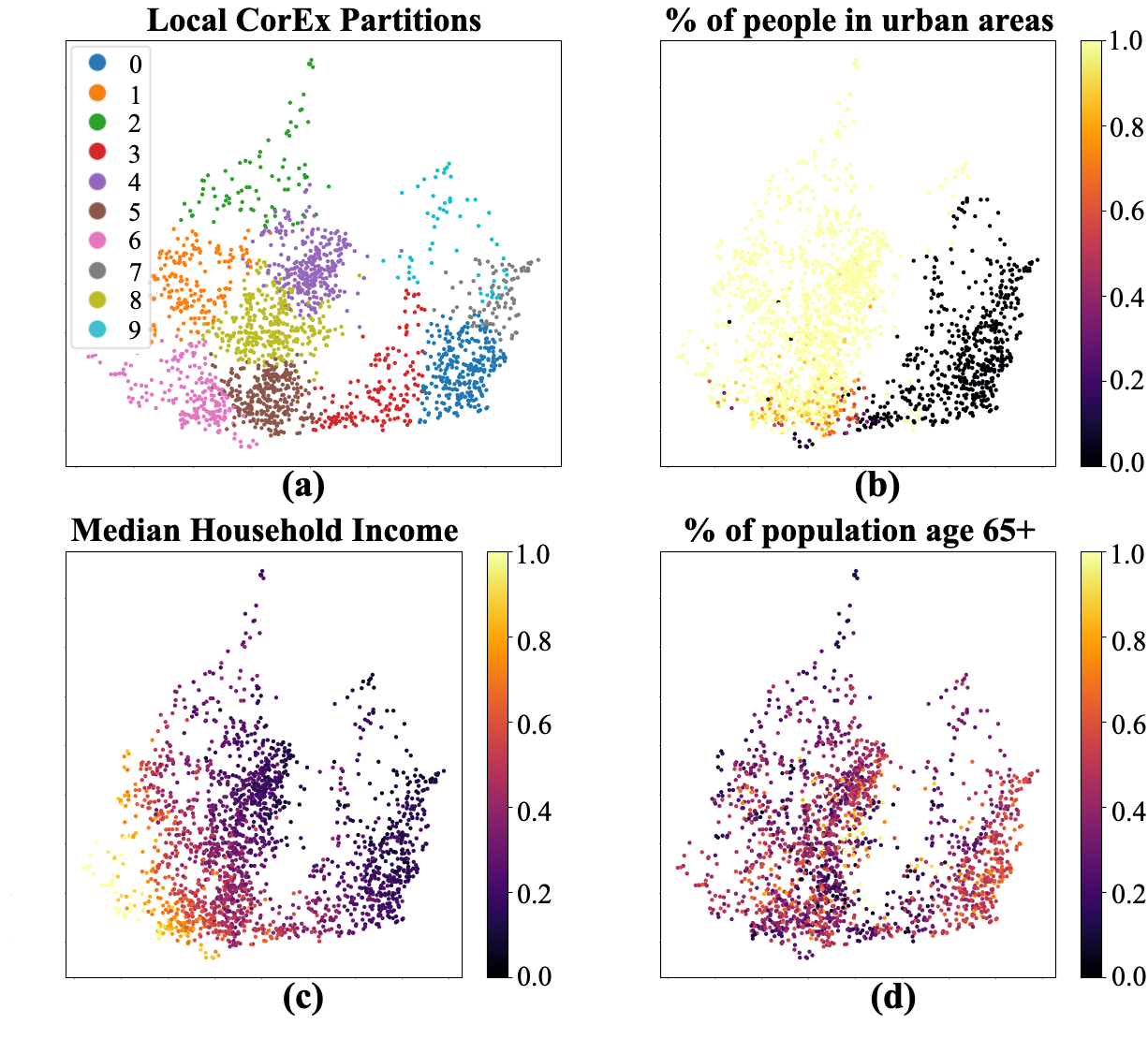}
            \caption{\textit{(a)} Two-dimensional PHATE embedding of the communities and crime dataset~\cite{misc_communities_and_crime_183} with the $10$ clusters resulting from Local CorEx plotted in different colors. \textit{(b)} PHATE embedding with the \% of people in urban areas colored. \textit{(c)} PHATE embedding with the median household income colored. \textit{(d)} PHATE embedding with the \% of the pop age 65+ colored.}
            \label{fig:crime_partitions}
        \end{figure}

        We remove all columns that contain missing values. Additionally, before partitioning the data we remove the community name, state, and cross-validation fold number columns. Next, we create a 12-dimensional PHATE embedding of the data and use $k$-means clustering with $k=10$. A 2D PHATE embedding of the chosen partitions is given in Figure \ref{fig:crime_partitions} as well as other PHATE plots colored by the percentage of people living in urban areas, the median household income, and the percentage of the population that is 65 or older. All variables are standardized between $0$ and $1$ even when some of the variable names imply a percentage. These plots show that the PHATE embedding captures meaningful structure in the data. The big cluster on the left is generally urban while the smaller cluster on the right is mostly rural. The median income increases the closer you get to the bottom left corner of the plot, and for each cluster, the lower right portion has some of the older populations. For this analysis, we further analyze clusters $0$ and $8$.

        Cluster 0 contains data for 282 communities. The top 5 states where these communities are based are Texas, Ohio, Oklahoma, Missouri, and New York with 30, 28, 18, 15, and 14 communities, respectively. From Figure \ref{fig:crime_partitions} we see that this cluster is rural, on the lower end for median income, and hosts an older population. When we look at the top 8 factors and visualize which features each factor is related to in Figure \ref{fig:group_0_factors} in the Appendix, we see that the factors group variables together which make sense. For example, The first factor relates the values of unmarried young people with living in the same city since 1985. The second factor groups the different measurements for rent prices and owner-occupied housing variables. The third factor relates the total percentage of the population that is divorced with the specific values for males and females as well as being somewhat correlated with the percentage of kids in family housing with two parents. 

        Cluster 8 contains data for 327 communities. The top 5 states associated with its communities are Florida, New Jersey, Massachusetts, Texas, and Washington with 56, 33, 19, 19, and 18 communities, respectively. Looking at Figure~\ref{fig:crime_partitions} we see that this cluster is very urban, on the lower to middle end for median income, and is not consistent for age above 65. We plot the mutual information between the top 8 factors with the original features in Figure \ref{fig:group_8_factors}, which have substantial overlap with those found in cluster 0. 
        
        We believe that this is because these groups are similar to the non-disjoint data in the ablation study. That is, the overall HOIs do not change much between groups in this setting, but the semantics of what is correlated will change between groups. For instance, latent factor 1 of cluster 0 and latent factor 6 of cluster 8 overlap substantially. They both have high mutual information with agePct16t24, agePct12t29, agePct12t21, and MalePctNevMarr. What is informative is when we examine the other variables related to these factors. For cluster 0, this factor also correlates with PctSameCity85 (percent of people living in the same city as in 1985, 5 years before), PctSameHouse85 (percent of people living in the same house as in 1985, 5 years before), PctBSorMore (percentage of people 25 and over with a bachelors degree or higher education), PctHousOwnOcc (percent of households owner occupied), PctPersOwnOccup (percent of people in owner-occupied households), and PctEmplProfServ (percentage of people 16 and over who are employed in professional services). This is informing us about the population of these young people. They are living in the same home and often the same city, in a home where the owner lives, and are working jobs in professional services. 

        Cluster 8's latent variables show a different picture. Factor 6 in this group is also associated with PctPopUnderPov (percentage of people under the poverty level), PctHousOwnOcc (percent of households owner-occupied), PctPersOwnOccup (percent of people in owner-occupied households), PctImmigRecent (percentage of immigrants who immigrated within last 3 years), agePct65up (percentage of population that is 65 and over in age), and PctImmigRec5 (percentage of immigrants who immigrated within last 5 years). Thus this cluster seems to be comprised of young people who are likely living under the poverty line, renting, and have recently immigrated to the US in the last 5 years.

        % Should I add a figure just comparing group 0 factor 0 to group 8 factor 6 rather than showing all of the factors in the appendix?

        By using Local CorEx we were able to find the main HOIs. We also were able to compare similar factors between clusters to discover interesting details that separate the clusters. Other interactions could also be considered.

    \subsection{MNIST}
    \label{sub:MNIST}

         \begin{figure}
            \centering
            \includegraphics[width=.47\textwidth]{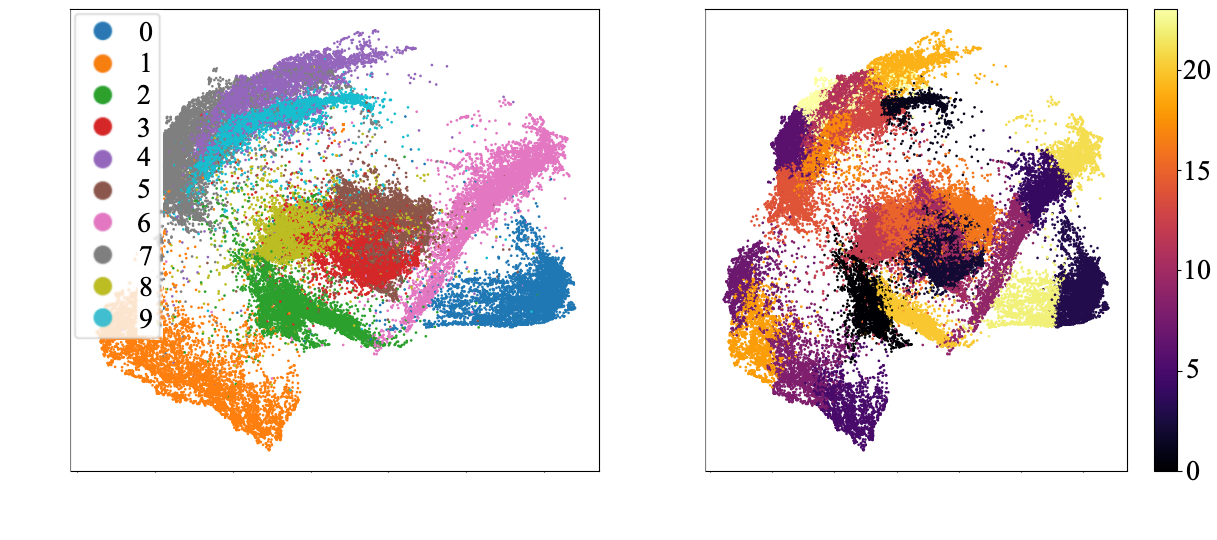}
            
            \includegraphics[width=.47\textwidth]{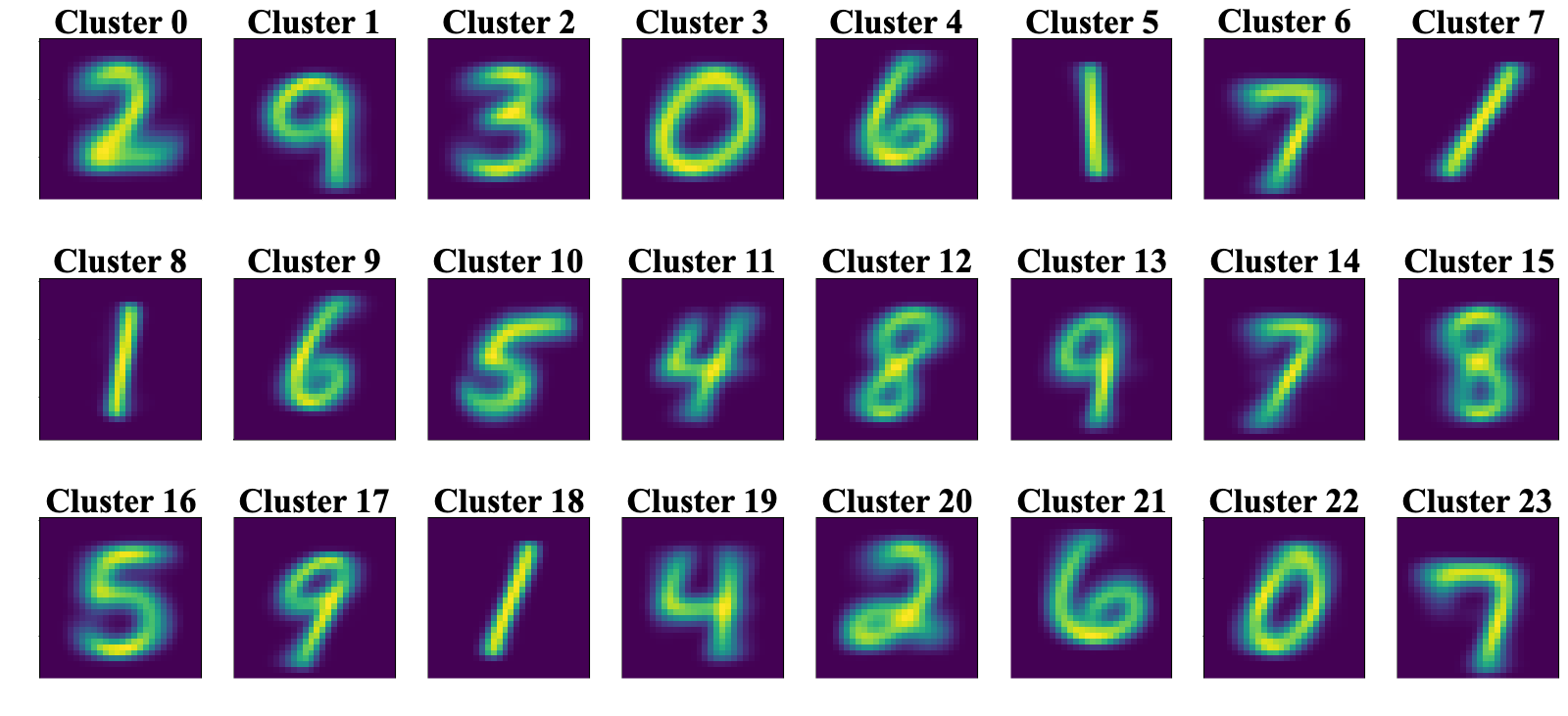}
            \caption{\textit{(Top left)} A two-dimensional PHATE embedding of the MNIST dataset colored by data labels. \textit{(Top right)} The same 2D PHATE embedding  colored by the Local CorEx clusters. The clusters largely respect the boundaries between classes. \textit{(Bottom)} The average pixel values found in each of the clusters. The outline of the majority digit present in each cluster is easily visible.}
            \label{fig:mnist_partitions}
        \end{figure}
        As another demonstration, we apply Local CorEx to the MNIST dataset \cite{lecun-mnisthandwrittendigit-2010}. The MNIST dataset is a common Machine Learning benchmark dataset for image classification. It consists of $70,000$ handwritten digits with $28 \times 28$ pixel images of centered digits. 

        We first partition the data by flattening the images into vectors of size $784$ and then creating a 10-dimensional PHATE embedding for the flattened data. We ran $k$-means clustering with $k=24$. The resulting clusters are given in Figure \ref{fig:mnist_partitions}.

\begin{figure}
            \centering
            \includegraphics[width=.47\textwidth]{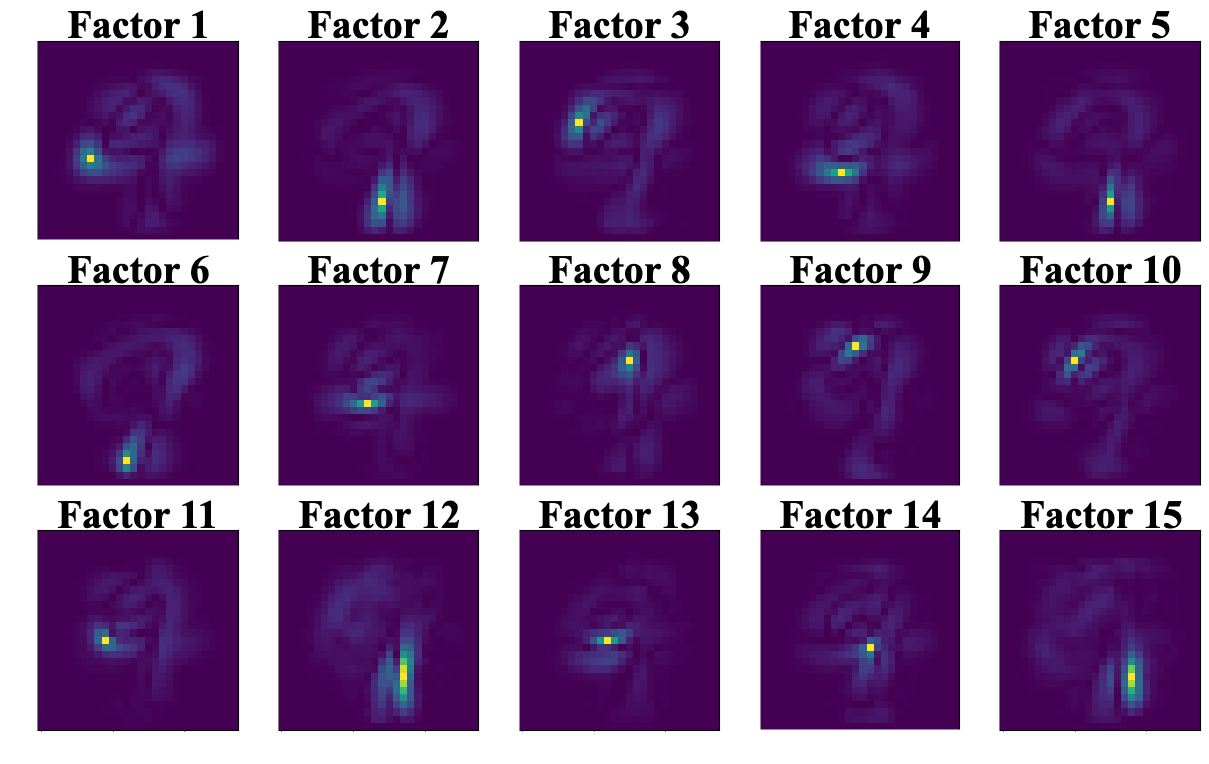}
            \caption{Visualizing the square-rooted mutual information between the first 15 Local CorEx factors trained on cluster 13 with the original features of the MNIST dataset.}
            \label{fig:group_13_corex_factors}
        \end{figure}

        We demonstrate Local CorEx using cluster 13, which has the following membership: 1898 9s, 1008 4s, 40 7s, 31 1s, 12 8s, and fewer than 10 each of the remaining digits. The square-rooted mutual information between each Local CorEx factor and the original features is shown in Figure \ref{fig:group_13_corex_factors}. Despite the data being extremely non-linear, Local CorEx is able to capture HOIs in the data that make sense. For example, factor 1 captures the far point where a four would likely bend while factor 2 captures the stem of the fours or nines. Other factors similarly appear to be related to constructing either fours or nines. While Local CorEx can be viewed as a piece-wise linear method, these results suggest that this is sufficient to capture many non-linearities.  See Appendix \ref{appendix_mnist_dataset} for additional examples of Local CorEx clusters.

     %   Although Local CorEx can be viewed as a piece-wise linear method we can see here that it is finding related variables for highly non-linear data. This is very encouraging and suggests that we might have faith that the method will also work for other non-linear data. See Appendix \ref{appendix_mnist_dataset} for additional examples of Local CorEx applied to other partitions. 
        
    \subsection{Neural Network Model Interpretability} \label{sub:neural_net}

        For our final case study, we use Local CorEx to explore the hidden representations and model weights for a neural network classifier trained on the MNIST dataset. After the classifier is fully trained, we train a decoder that is used to visualize reconstructed perturbed hidden layer representations. The full model architecture is in Figure \ref{fig:model_architectures}. All explorations are conducted on the test data which is partitioned into 20 clusters using $k$-means clustering on a 10-dimensional PHATE embedding of the data. The average pixel value in each cluster is visualized in Figure \ref{fig:mnist_test_partitions}. For the following analyses, we focus on cluster 16 for brevity but analyze an additional cluster in Appendix \ref{neural_network}. Cluster 16 was chosen due to the heterogeneity of the class labels between 9s and 4s and its similarities to cluster 13 in Section \ref{sub:MNIST}.
        
        \subsubsection{Effect of Dropout}
        It is assumed that neural networks trained with dropout \cite{dropout_2014} tend to duplicate the passing of information across layers, so when a node is dropped the information is still passed forward. If this is true, then the duplication of information passed would also increase the total correlation present in a layer. To explore this, we trained a series of paired classifier models on the MNIST dataset. All models have two hidden layers and share all hyperparameters except, in each pair, one uses dropout with $0.5$ and the other does not use dropout. To help simplify the comparison, each pair of models uses the same initialized set of weights.

        We feed cluster 16 into the trained classifiers, where the output of each layer is viewed separately as a hidden state representation of the data. We then run Linear CorEx on each of the hidden states (i.e. layer outputs) separately, measure the total correlation explained by each latent factor, and report the summed total correlation values for the first 5 latent factors. These total correlation measurements across hidden layers should measure the total amount of information shared among the hidden nodes and be a good surrogate for how much redundancy of information is present in a hidden layer. We repeated this study on pairs of classifier models with 3 hidden layers to further explore the effect of deeper networks. Table \ref{tab:tc_dropout} shows the average values obtained across the 10 pairs of models for both 2 and 3 hidden layers.

        % I'm not sure if I should report TC and L1-norm values or just one or the other.
        \begin{table}[t]
            \caption{Average Total Correlation values of the mutual information between the first 5 CorEx factors and the original features. Total correlation increases as models become deeper both with and without dropout. Dropout exacerbates this trend in the later layers.}
            \label{tab:tc_dropout}
            \begin{center}
            \begin{small}
            \begin{sc}
            \begin{tabular}{lrr}
            \toprule
            & \multicolumn{2}{c}{Total Correlation}\\
            Layer & no dropout & with dropout \\
            \midrule
            2 layer H1 & 31.2 $\pm$ 1.6 & 27.2 $\pm$ 2.0 \\
            2 layer H2 & 51.9 $\pm$ 2.8 & 98.1 $\pm$ 4.4 \\
            3 layer H1 & 33.7 $\pm$ 2.5 & 26.4 $\pm$ 1.2 \\
            3 layer H2 & 40.8 $\pm$ 2.8 & 95.9 $\pm$ 3.2 \\
            3 layer H3 & 76.1 $\pm$ 4.6 & 162.4 $\pm$ 8.5 \\
            \bottomrule
            \end{tabular}
            \end{sc}
            \end{small}
            \end{center}
        \end{table}

        % Does this paragraph make sense?
        The total correlation in the first hidden layer is slightly smaller for the models with dropout. This is in stark contrast to what we see for the later layers where the amount of total correlation is substantially larger in the models with dropout compared to the models without. Our hypothesis for this is that the first layer extracts basic features from the data, such as relevant clusters of pixels. These features are by nature sparse and likely do not benefit much from the dropout push to become less sparse. In later layers, the model has had the chance to learn useful features that it can distribute across multiple nodes to ensure that the feature is preserved if the node is dropped. This suggests that the benefits of dropout may not be equal in different layers within the neural network. %This is why we believe that the total correlation in the first layer is virtually the same between models with 2 and 3 layers while the total correlation in the second layer differs between models with 2 and 3 hidden layers.

        \subsubsection{Visualizing hidden node effects}\label{subsub:vis_hidden_node}

          \begin{figure*}
            \centering
            \includegraphics[width=.48\linewidth]{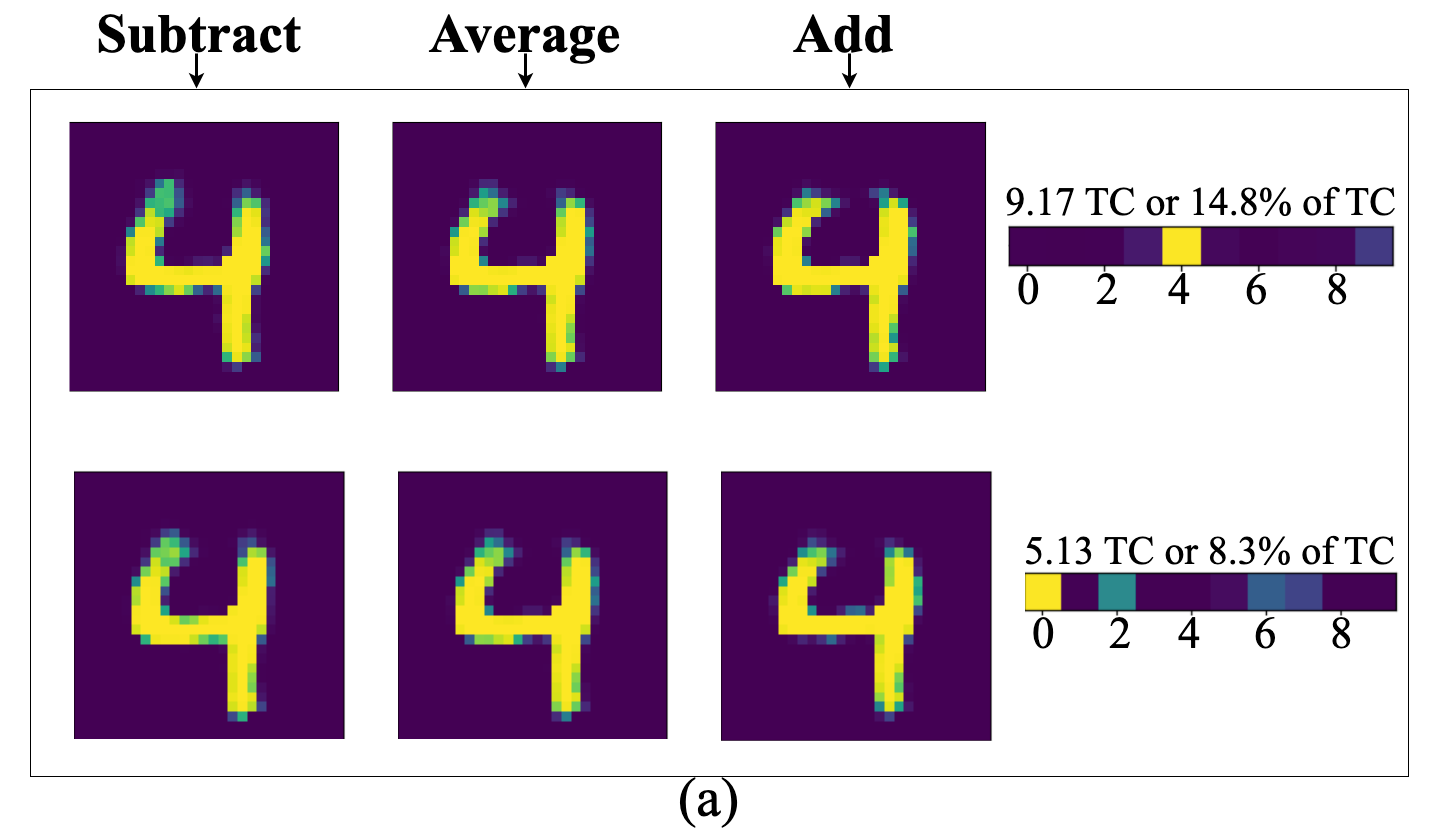}
            \includegraphics[width=.48\linewidth]{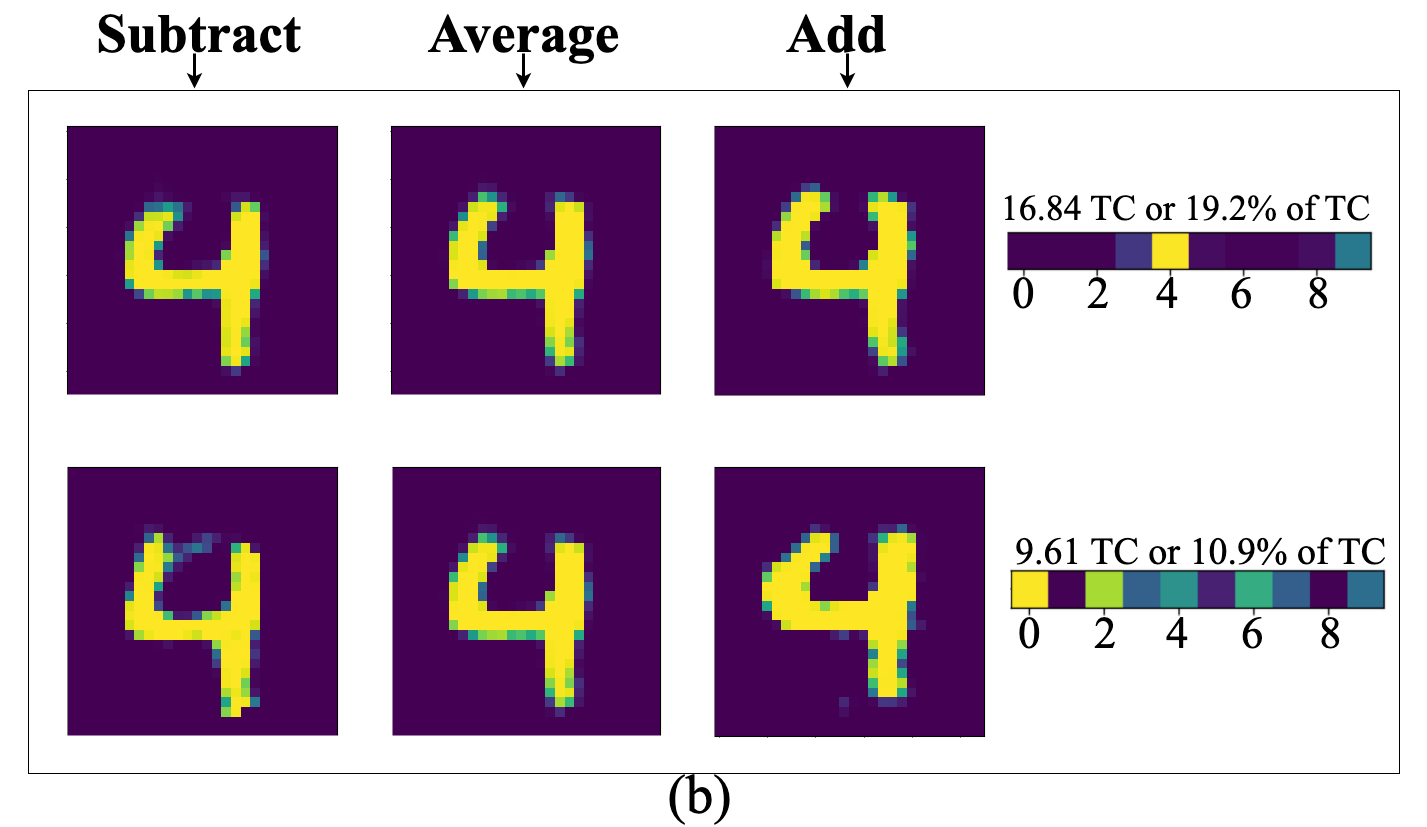}
            \caption{Visualizing the effect of perturbing the average neural network hidden state representations of cluster 16 in the MNIST test dataset. \textit{(a)} The plots are associated with perturbing the H1 representation. This first row is associated with the first Local CorEx factor and the second row is associated with the second Local CorEx factor. \textit{(b)} Same as in \textit{(a)} but for H2.  For each group of plots, the leftmost column image is generated by subtracting the mutual information between the Local CorEx factor and the hidden nodes from the average representation. The second column image gives the average hidden state representation. The third column image is generated by adding the mutual information between the Local CorEx factor and the hidden nodes from the average representation. Finally, the rightmost column plots the mutual information between the Local CorEx factor and the model logits. This analysis gives us a visual intuition for what role the grouped hidden nodes play.}
            \label{fig:reconstructed_hs}
        \end{figure*}

        Here we analyze the effects of nodes in the hidden layer of a neural network by running Local CorEx on the concatenation of a hidden layer output with their respective logits (see Figure \ref{fig:neural_net_processing}). The logits are included to aid with interpretability. Using this we can find sets of related hidden nodes and get an intuition for which class they're associated with using the logits. To visualize their effect we trained a decoder and used the first two layers of the classifier as an encoder with frozen weights. This allows us to transform hidden state representations into reconstructed inputs. 

        Using this setup, we took the average first hidden layer representation for cluster 16 and perturbed the hidden nodes found by the Local CorEx factor in proportion to their mutual information with the hidden nodes and then decoded the resulting perturbed first hidden layer representation. The results can be seen for the first 2 latent factors on the left side of Figure \ref{fig:reconstructed_hs}. Perturbing the values of the hidden nodes associated with the first latent factor alters the curvature of the digit in the top left portion of the number. Perturbing the hidden nodes associated with the second Local CorEx factor removes pixels on the base of the four, the base of the horizontal line in the center, and the top part of the number.  Removing the pixels in these locations is associated with the probability of the sample being classified as a zero or a two, as indicated by the mutual information.

        We repeat this for the second hidden layer, with the results shown on the right side of Figure \ref{fig:reconstructed_hs}. Here we see very similar effects as that of perturbing the first hidden layer. Perturbing the hidden nodes associated with the first Local CorEx factor changes how the pixels come together or not near the top of the four. This factor is mostly associated with the 4 and 9 logits. The second Local CorEx factor affects the digit's thickness, the location of the middle bar, the angle of the top left line, and where the base of the stem begins. This factor is associated with various logits that will be influenced by these pixels.  %Given the fact that we have access to the sample labels, we could also replace the logits with a one-hot encoding of the labels to better understand the differences between two different classes that appear close to one another in the input space.

        \subsubsection{Quantifying hidden node effect on classifier performance} \label{subsub:quantify_node_effect}

        In the previous section, we visually inspected the effect of perturbing hidden nodes associated with a Local CorEx factor. Here we delete the hidden nodes in the classifier model and measure their impact on the classifier's performance across clusters to determine if the related nodes have a local or global impact on the classifier's performance. We start by obtaining a baseline to see how the classifier performs across clusters, with the results shown in Table \ref{tab:mnist_classifer_performance} in the Appendix. The model performs relatively well across all clusters. We then use the same Local CorEx factors learned in Section \ref{subsub:vis_hidden_node} and create reduced models by deleting the hidden nodes associated with each of the factors. We then recompute the accuracies across all clusters using the reduced model. 
        
\begin{figure}
            \centering
            \includegraphics[width=.48\textwidth]{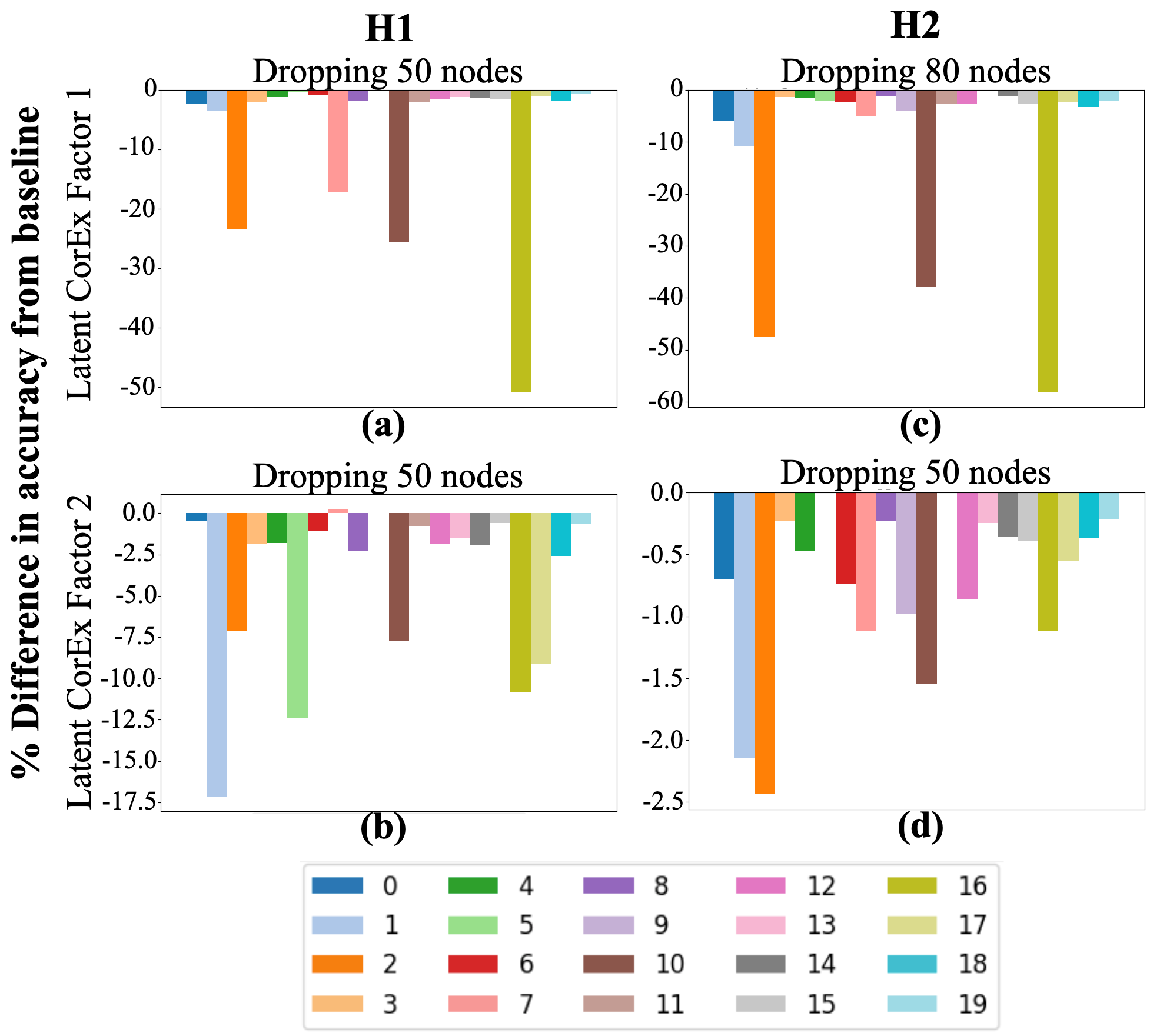}
            \caption{\textit{(a)} Difference in classification accuracy between the unaltered model and the model after deleting 50 hidden nodes in the first hidden layer with the highest mutual information with the first Local CorEx factor associated with cluster 16. \textit(b) Same as \textit{(a)} except when using the second CorEx factor to determine the 50 hidden nodes to delete.  \textit(c) Same as in \textit{(a)} except when deleting 80 hidden nodes in the second hidden layer.  \textit(d) Same as in \textit{(b)} except in the second hidden layer.  Note that the y-axis scale differs for each plot.} %The first CorEx factors for both H1 and H2 are associated with the logit for classifying fours. The groups of nodes associated with these factors are critical for accurately classifying the digits in clusters 2, 10, and 16 which contain mainly 4s and 9s. The second CorEx factors for both H1 and H2 correlate with logits for digits 0, 2, and 6 and seem to encode information less necessary for accurately classifying one particular group.}
            \label{fig:dropping_hidden_nodes}
        \end{figure}

        We start with the Local CorEx factors learned from the H1 embeddings concatenated with logits. The first factor is associated with the logit for classifying fours. When we delete the 50 nodes with the highest mutual information associated with this factor and recompute the accuracies across clusters, it has a large impact on classification accuracy for clusters 2, 10, and 16, as shown in Figure~\ref{fig:dropping_hidden_nodes}. All of these clusters have large quantities of 4s and 9s present. Remarkably, almost all other clusters are relatively unaffected by this despite deleting 25\% of the hidden nodes in the first layer. The second factor is associated with several logits including 0's, 2's, and 6's. When 25\% of the nodes associated with this factor are dropped the effect is more widespread. The clusters most impacted are 1, 2, 5, 10, 16, and 17. Clusters 2, 10, and 16 contain 4s and 9s so it isn't surprising that removing a feature calculated on a group of 4s and 9s affects their classification. Groups 1, 5, and 17 are composed almost entirely of 6s. So although this factor has high mutual information with the 0, 2, and 6 logits it affects mostly the classification accuracy of groups of 6s.

        We continue our analysis with the H2 embeddings concatenated with logits. The first Local CorEx factor is associated with the logit for classifying fours. When we delete the first 80 nodes with the highest mutual information associated with it and recompute the accuracies across clusters we see it has a large impact on classification accuracy for clusters 2, 10, and 16 which contain 4s and 9s. Almost all other clusters are relatively unaffected by this despite deleting 40\% of the hidden nodes in the second hidden layer. Repeating this for the second latent factor, which is associated with several logits including 0's, 2's, and 6's, seems to have little effect on the model's overall performance despite it representing 25\% of the layer's hidden nodes. The clusters most affected though are 1, 2, 10, and 16, which contain mostly digits with a flat horizontal line in the center of the digit (Figure \ref{fig:mnist_test_partitions}). Additionally when we look at the effect of perturbing this factor in Figure \ref{fig:reconstructed_hs} we see that it affects the height of the middle bar. Although this feature is less important for classification, the effect on accuracy across partitions makes sense. 

        % Add Main Figure explaining how neural net analysis is conducted.

\section{Conclusion} \label{Discussion}
    % Mention it is interesting for the neural network interpretability that we are leveraging 3 representations of the data and in the other case studies we leverage 2 representations. We could generalize this method to simply calculate the M.I. between any two data representations

    %One of the reasons that we believe this method works is the following statement that we believe to be true. If a HOI is an ``important" component of the data structure then a ``good" representation of the data will preserve that interaction. Assuming that the second representation has a different basis than the first we can compare the two representations to try to discern what information is preserved and how. CorEx creates a latent factor representation where we can view how its features relate to the original feature space by looking at the mutual information between the two. We believe this approach can be expanded to be used with any two representations where you have some way to compare the two. % This is obviously too vague, but I wanted to write down the gist before I forgot about it.

%\section{Conclusion} \label{Conclusion}

We have shown that Local CorEx is superior to global methods at capturing variable interactions when these interactions vary across the data manifold. In Table \ref{tab:full_ablation_study}, we demonstrated that even when interactions largely overlap, Local CorEx outperforms global methods with mostly pure clusters when the sample size exceeds 100. Furthermore, despite impure clusters, Local CorEx performs comparably to global methods once the sample size reaches 1000. The more the variable interactions vary, the greater the benefit of using Local CorEx over other global methods for extracting HOIs in the data. 

We then effectively used Local CorEx to explore several different data types to extract meaningful variable interactions, including tabular data, image data, neural network nodes, and internal representations. With the neural network, we used Local CorEx in an unsupervised manner to determine the hidden nodes the model leveraged for predicting a specific class and showed that by removing them only that main class was affected. This demonstrates that despite the interconnectedness of neural networks we can isolate clusters of nodes that perform a particular task of interest. We also confirmed that using dropout in a neural network increases the redundancy or total correlation present in the representation of the layer in the later layers of the network, but not in the first layer.

We believe that this approach can be used to further explore and interpret the inner workings of neural networks. For example, Local CorEx could be used to study robustness in neural networks by identifying features the network associates with a class and use them to create adversarial data points to improve model performance in a manner conceptually similar to work done by \cite{casper_2022,casper2023diagnostics}. Additionally, Local CorEx can be used as part of exploratory data analysis to detect variable interactions in different regions of the data manifold such as in ecological systems \cite{grilli2017higher, levine_2022}, collaborations \cite{civilini2023explosive}, the human brain \cite{petri2014homological, giusti2016two}, and any network-based data including biological networks. For future work, we believe this method can be further adapted as a visualization tool to aid in summarizing complex datasets.


\section*{Acknowledgments}
This research was supported in part by the NSF under Grant 2212325. The content provided here is solely the responsibility of the authors and does not necessarily represent the official views of the funding agency.

% \section{Impact Statement}

% This paper presents work whose goal is to advance the field of Machine Learning. There are many potential societal consequences of our work, none which we feel must be specifically highlighted here.

\bibliography{main}
\bibliographystyle{icml2024}

%%%%%%%%%%%%%%%%%%%%%%%%%%%%%%%%%%%%%%%%%%%%%%%%%%%%%%%%%%%%%%%%%%%%%%%%%%%%%%%
%%%%%%%%%%%%%%%%%%%%%%%%%%%%%%%%%%%%%%%%%%%%%%%%%%%%%%%%%%%%%%%%%%%%%%%%%%%%%%%
% APPENDIX
%%%%%%%%%%%%%%%%%%%%%%%%%%%%%%%%%%%%%%%%%%%%%%%%%%%%%%%%%%%%%%%%%%%%%%%%%%%%%%%
%%%%%%%%%%%%%%%%%%%%%%%%%%%%%%%%%%%%%%%%%%%%%%%%%%%%%%%%%%%%%%%%%%%%%%%%%%%%%%%
\newpage
\appendix
\onecolumn
\section{Ablation Study Details} \label{appendix_ablation_study}

\begin{table*}[t]
    \caption{Full ablation study results.}
    \label{tab:full_ablation_study}
    \begin{center}
    \begin{small}
    \begin{sc}
    \begin{tabular}{lllrrrrrrrr}
    \toprule
     & & & \multicolumn{4}{c}{Group} & \multicolumn{4}{c}{Top K Latent Factor}\\
    & & & \multicolumn{2}{c}{Cosine Dist} & \multicolumn{2}{c}{AUCPRC} & \multicolumn{2}{c}{Cosine Dist} & \multicolumn{2}{c}{AUCPRC}\\
    Data & Size & Method & $\alpha=0.0$ & $\alpha=1.0$ & $\alpha=0.0$ & $\alpha=1.0$ & $\alpha=0.0$ & $\alpha=1.0$ & $\alpha=0.0$ & $\alpha=1.0$ \\
    \midrule
    Disjoint & 10 & Linear & 0.514 & 0.489 & \textbf{0.529} & 0.505 & \textbf{0.464} & 0.451 & \textbf{0.581} & 0.548 \\
     &  & Bio & 0.541 & 0.455 & 0.475 & 0.488 & 0.500 & 0.454 & 0.522 & 0.545 \\
     &  & Local Lin & 0.539 & 0.413 & 0.477 & \textbf{0.663} & 0.481 & \textbf{0.352} & 0.517 & \textbf{0.730} \\
     &  & Local Bio & \textbf{0.359} & \textbf{0.394} & 0.515 & 0.557 & 0.485 & 0.434 & 0.523 & 0.600 \\
     \cline{2-11}\\
     & 100 & Linear & 0.484 & 0.424 & 0.607 & 0.591 & 0.406 & 0.403 & 0.702 & 0.627 \\
     &  & Bio & \textbf{0.447} & 0.473 & 0.607 & 0.509 & \textbf{0.351} & 0.436 & \textbf{0.716} & 0.602 \\
     &  & Local Lin & 0.477 &\textbf{ 0.190} & \textbf{0.610} & \textbf{0.894} & 0.408 & 0.086 & 0.687 & \textbf{0.989} \\
     &  & Local Bio & 0.473 & 0.226 & 0.563 & 0.841 & 0.388 & \textbf{0.080} & 0.654 & 0.980 \\
     \cline{2-11}\\
     & 1000 & Linear & 0.516 & 0.408 & 0.606 & 0.604 & 0.412 & 0.401 & 0.701 & 0.640 \\
     &  & Bio & \textbf{0.414} & 0.474 & \textbf{0.652} & 0.507 & \textbf{0.323} & 0.436 & \textbf{0.758} & 0.617 \\
     &  & Local Lin & 0.504 & \textbf{0.146} & 0.607 & \textbf{0.902} & 0.420 & 0.031 & 0.691 & 0.999 \\
     &  & Local Bio & 0.451 & 0.163 & 0.610 & 0.883 & 0.374 & \textbf{0.028} & 0.708 & \textbf{1.000} \\
     \cline{2-11}\\
     & 10000 & Linear & 0.526 & 0.412 & 0.599 & 0.598 & 0.411 & 0.408 & 0.696 & 0.633 \\
     &  & Local Lin & \textbf{0.140} & \textbf{0.142} & \textbf{0.907} & \textbf{0.907} & \textbf{0.024} & \textbf{0.026} & \textbf{1.000} & \textbf{1.000} \\
     \cline{1-11}\\
     
    Non- & 10 & Linear & \textbf{0.354} & \textbf{0.255} & \textbf{0.739} & \textbf{0.801} & \textbf{0.278} & \textbf{0.197} & \textbf{0.828} & \textbf{0.871} \\
    disjoint &  & Bio  & 0.401 & 0.312 & 0.634 & 0.670 & 0.308 & 0.290 & 0.751 & 0.763 \\
     &  & Local Lin & 0.453 & 0.385 & 0.626 & 0.712 & 0.402 & 0.341 & 0.697 & 0.770 \\
     &  & Local Bio    & 0.371 & 0.377 & 0.569 & 0.585 & 0.443 & 0.410 & 0.624 & 0.638 \\
     \cline{2-11}\\
     & 100 & Linear    & \textbf{0.207} & \textbf{0.197} & \textbf{0.883} & 0.856 & \textbf{0.107} & 0.103 & \textbf{0.989} & 0.952 \\
     &  & Bio          & 0.236 & 0.289 & 0.842 & 0.754 & \textbf{0.107} & 0.206 & 0.981 & 0.888 \\
     &  & Local Lin & 0.254 & \textbf{0.197} & 0.844 & \textbf{0.890} & 0.163 & 0.106 & 0.934 & \textbf{0.987} \\
     &  & Local Bio    & 0.289 & 0.239 & 0.782 & 0.827 & 0.157 & \textbf{0.088} & 0.920 & 0.979 \\
     \cline{2-11}\\
     & 1000 & Linear   & \textbf{0.191} & 0.182 & \textbf{0.886} & 0.856 & 0.091 & 0.086 & \textbf{0.992} & 0.959 \\
     &  & Bio          & 0.203 & 0.276 & 0.869 & 0.773 & \textbf{0.084} & 0.199 & 0.991 & 0.892 \\
     &  & Local Lin & 0.197 & \textbf{0.141} & 0.870 & \textbf{0.908} & 0.100 & 0.051 & 0.959 & \textbf{1.000} \\
     &  & Local Bio    & 0.222 & 0.173 & 0.850 & 0.873 & 0.107 & \textbf{0.039} & 0.965 & \textbf{1.000} \\
     \cline{2-11}\\
     & 10000 & Linear  & \textbf{0.189} & 0.183 & \textbf{0.887} & 0.855 & \textbf{0.089} & 0.086 & \textbf{0.993} & 0.958 \\
     &  & Local Lin & 0.190 & \textbf{0.139} & 0.883 & \textbf{0.912} & 0.090 & \textbf{0.048} & \textbf{0.993} & \textbf{1.000} \\
    \bottomrule
    \end{tabular}
    \end{sc}
    \end{small}
    \end{center}
\end{table*}

\begin{algorithm}[tb]
   \caption{Group scoring}
   \label{alg:group_scoring}
\begin{algorithmic}
   \STATE {\bfseries Input:} CorEx Factors $F$, HOIs $G$
   \STATE Initialize $cos\_score = empty\_list$, $aucprc\_score = empty\_list$, $k = $ size of $G$.
   \FOR{$g$ {\bfseries in} $G$}
   \STATE Initialize $best\_cosine = 1$, $best\_aucprc = 0$.
   \FOR{$f$ {\bfseries in} $F$}
   \STATE $cos = cosine\_distance(f, g)$
   \STATE $aucprc = auc\_prc(f, g)$
   \IF{$cos < best\_cosine$}
   \STATE Update $best\_cosine$
   \ENDIF
   \IF{$aucprc > best\_aucprc$}
   \STATE Update $best\_aucprc$
   \ENDIF
   \ENDFOR
   \STATE Append $best\_cosine$ to $cos\_score$
   \STATE Append $best\_aucprc$ to $aucprc\_score$
   \ENDFOR \\
   \textbf{Return} Average $cos\_score$ and $aucprc\_score$
\end{algorithmic}
\end{algorithm}

\begin{algorithm}[tb]
   \caption{Top-k scoring}
   \label{alg:top_k_scoring}
\begin{algorithmic}
   \STATE {\bfseries Input:} CorEx Factors $F$, HOIs $G$
   \STATE Initialize $cos\_score = empty\_list$, $aucprc\_score = empty\_list$, $k = $ size of $G$.
   \IF{size of $G > $ size of $F$}
   \STATE Update $k = $ size of $F$
   \ENDIF
   \FOR{$i=1$ {\bfseries to} $k$}
   \STATE Initialize $best\_cosine = 1$, $best\_aucprc = 0$.
   \FOR{$g$ {\bfseries in} $G$}
   \STATE $cos = cosine\_distance(f, g)$
   \STATE $aucprc = auc\_prc(f, g)$
   \IF{$cos < best\_cosine$}
   \STATE Update $best\_cosine$
   \ENDIF
   \IF{$aucprc > best\_aucprc$}
   \STATE Update $best\_aucprc$
   \ENDIF
   \ENDFOR
   \STATE Append $best\_cosine$ to $cos\_score$
   \STATE Append $best\_aucprc$ to $aucprc\_score$
   \ENDFOR \\
   \textbf{Return} Average $cos\_score$ and $aucprc\_score$
\end{algorithmic}
\end{algorithm}

\begin{figure}
    \centering
    \includegraphics[width=.3\textwidth]{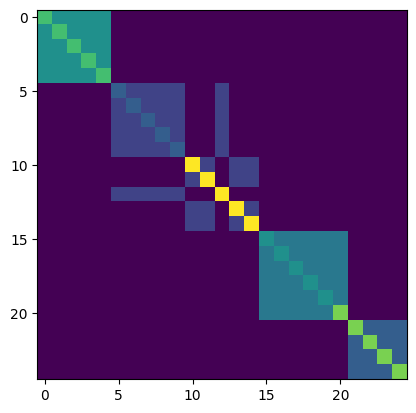}
    \includegraphics[width=.3\textwidth]{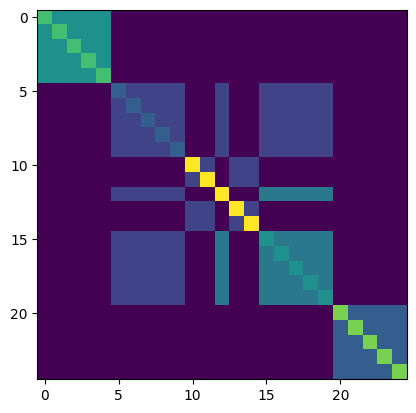}
    \includegraphics[width=.3\textwidth]{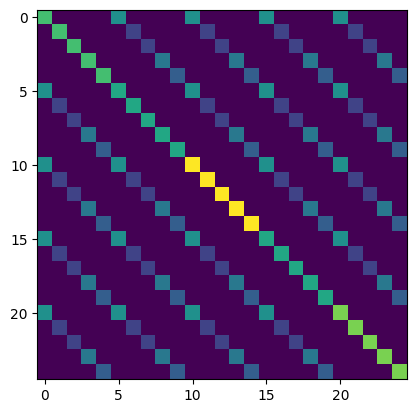}
    \caption{The set of covariance matrices used in generating simulated data for the ablation study. The matrix on the left was the covariance matrix used in all simulations for Cluster 1. Cluster 2 uses either the middle or right matrix for its covariance matrix depending on whether we were simulating non-disjoint (middle) or disjoint (right) HOIs between the two clusters.}
    \label{fig:covariance_matrices}
\end{figure}

Here we describe the ablation study in Section~\ref{sub:ablation} in more detail. Figure \ref{fig:covariance_matrices} shows the set of covariance matrices used in the ablation study synthetic data. The first multivariate normal distribution always uses the far left covariance matrix, while the second distribution uses the middle and left covariance matrices for different replications to simulate when the clusters' higher order interactions stayed mostly the same (non-disjoint) or were completely different (disjoint), respectively.

% FIXME need to add axis to these matrices so they know the values there

The means of each cluster are 
\begin{center}
    $\mu_1 = (5,5,5,5,5,10,10,10,10,10,1,1,1,1,1,10,10,10,10,10,3,3,3,3,3)$ \\
    $\mu_2 = \mu_1 + \alpha * (2.5, 2.5, 2.5, 2.5, 2.5, 4, 4, 4, 4, 4, -2, -2, -2, -2, -2, 9, 9, 9, 9, 9, -7, -7, -7, -7, -7),$
\end{center} 
 where $\alpha$ varies linearly between $0$ and $1$ in increments of $0.1$. For each simulation setup we ran the CorEx algorithms using 3 through 7 latent factors so we could see the effect of having too few and too many factors (there were 5 true HOIs). The effect of the number of latent factors used on the scores calculated using Algorithms \ref{alg:group_scoring} \& \ref{alg:top_k_scoring} was minimal so all reported metrics are averaged across the different latent factors.

\begin{figure}
    \centering
    \includegraphics[width=0.87\textwidth]{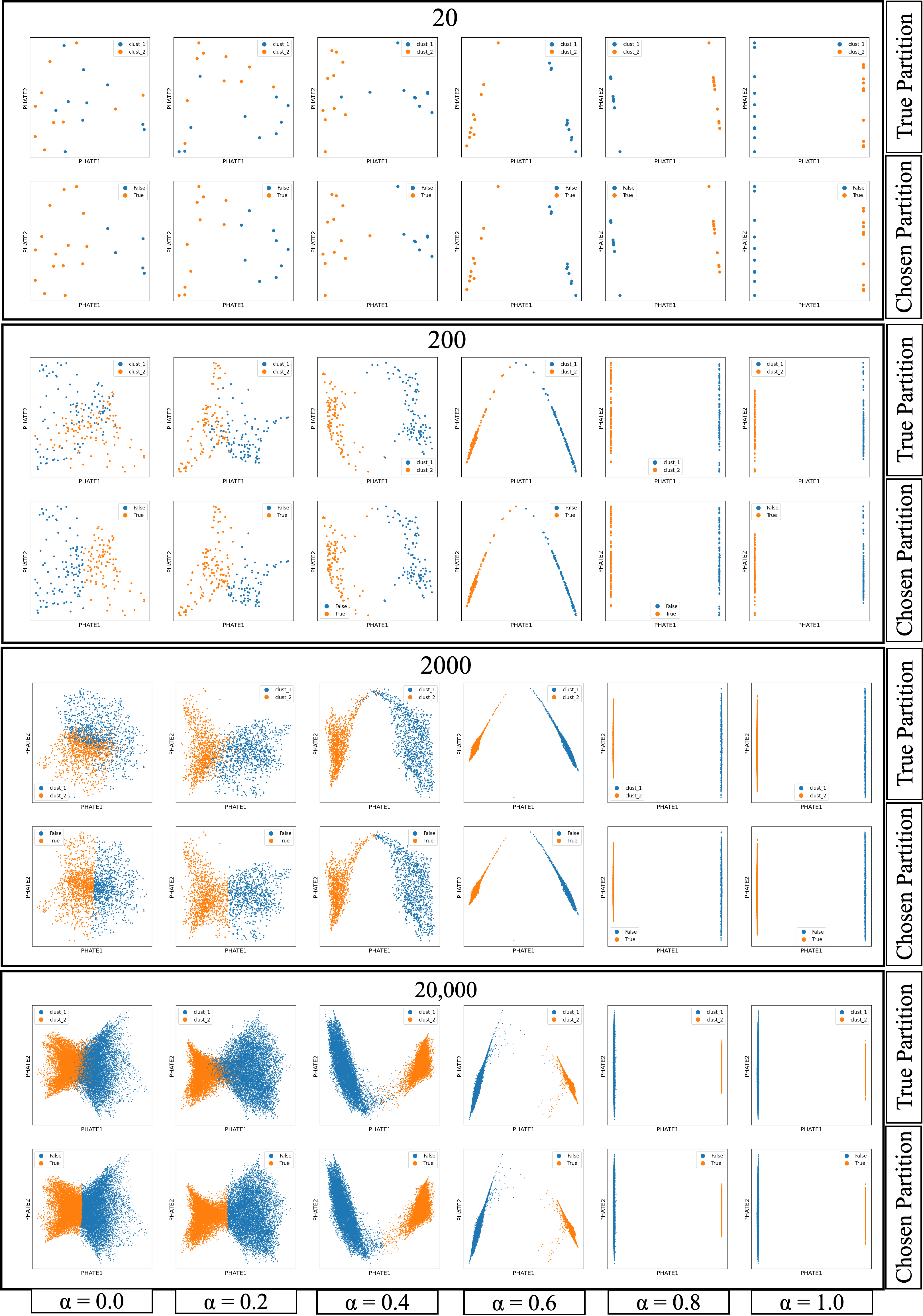}
    \caption{The x-axis varies the $\alpha$ parameter in the synthetic data starting at 0.0 on the left and increasing in increments of 0.1 until 1.0 on the far right. The four sets of plots increase in sample size starting at 20 on the top and continuing to 200, 2,000, and 20,000 on the bottom. For each set, the top plot shows the true partitions and the bottom plot shows the chosen partition.}
    \label{fig:sample_size_alphas}
\end{figure}

\begin{figure}
    \centering
    \includegraphics[width=1.0\textwidth]{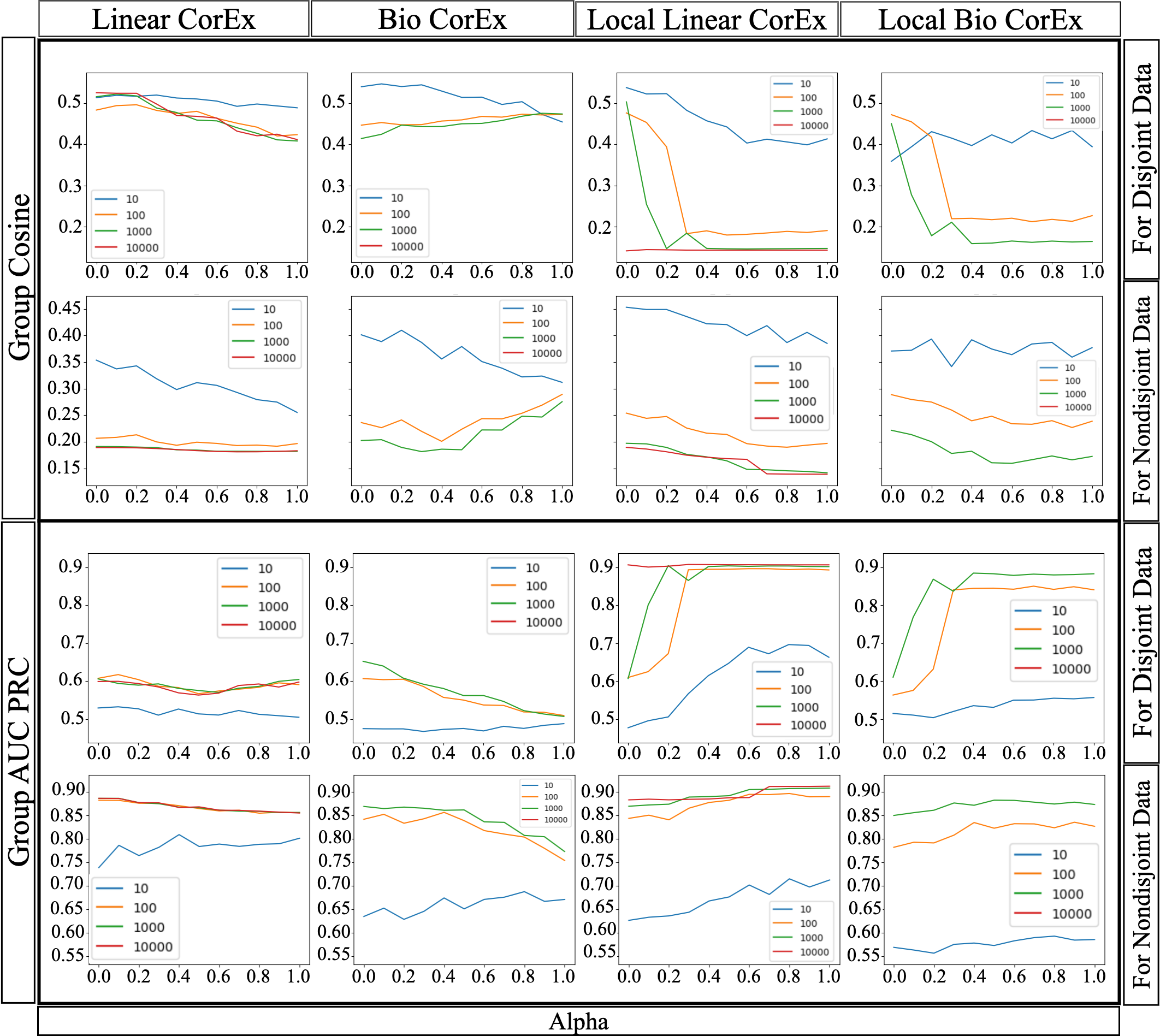}
    \caption{The group results from the synthetic data. The leftmost plots show results from the Linear CorEx method applied to all the data and the second column is the same except using the Bio CorEx algorithm. The 3rd and 4th columns use Local CorEx applying the Linear and Bio CorEx algorithms to the clusters, respectively. All plots are colored based on the sample size and the x-axis indicates the alpha level. \textit{(Top row)} The cosine distance between each ground truth interaction and the closest latent factor averaged across replicates and the number of latent factors for the disjoint data. \textit{(Second row)} Same as the first row except for the non-disjoint data. \textit{(Third row)} The average area under the precision-recall curve between each ground truth interaction and the closest latent factor averaged across replicates and the number of latent factors for the disjoint data. \textit{(Fourth row)} Same as the third row except for the non-disjoint data. Note that for disjoint data local methods receive a boost once the data becomes separable. Nondisjoint data also benefit from increased separability but the jump isn't as large. Additionally, this lets us know that overall we have good recall of all of the HOIs present with sufficient samples and good clustering.}
    \label{fig:group_scores}
\end{figure}

2D PHATE embeddings for a single simulation for the different sample sizes and alpha values are given in Figure \ref{fig:sample_size_alphas}. The results in Figure \ref{fig:group_scores} show that the local methods (3rd and 4th columns) outperform or match the performance of the global methods (1st and 2nd columns) except for the case when the sample size is 10 for each cluster. For the global partitions, the sample size is double that of the local. Thus when the interactions don't differ much between clusters and the sample size is small, the increase in sample size outweighs the effect of having pure clusters. Another takeaway is that despite Bio CorEx being a more flexible and non-linear method it does not perform better on average over Linear CorEx. 
        
\begin{figure}
    \centering
    \includegraphics[width=1.0\textwidth]{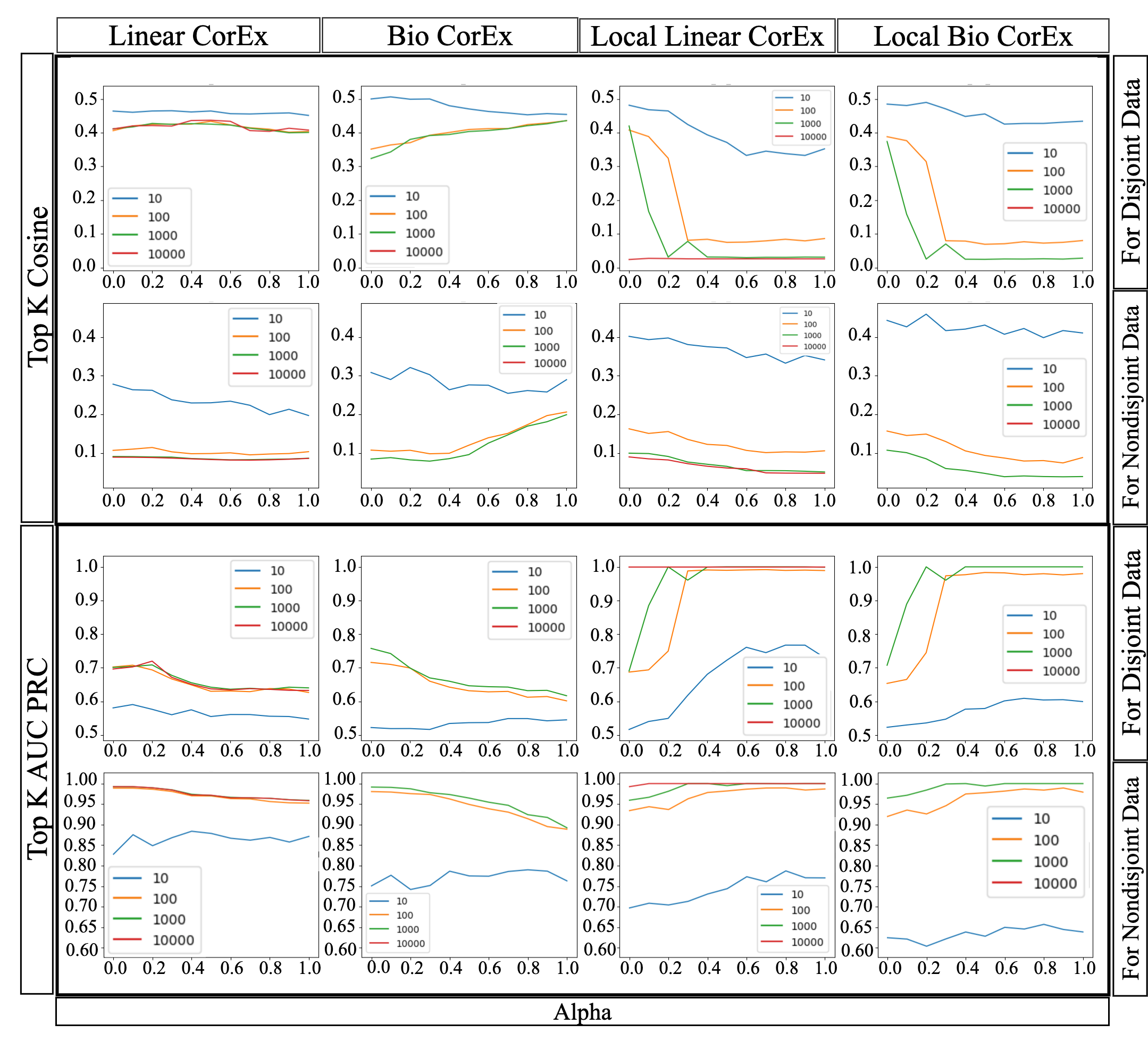}
    \caption{The top-$k$ results of the synthetic data. The leftmost plots show results from the Linear CorEx method applied to all the data and the second column is the same except using the Bio CorEx algorithm. The 3rd and 4th columns use Local CorEx applying the Linear and Bio CorEx algorithms to the clusters, respectively. All plots are colored based on the sample size and the x-axis indicates the alpha level. Top row: The cosine distance between each of the at-most $k$ interactions found using CorEx with the closest ground truth interaction averaged across replicates and the number of latent factors for the disjoint data. So if there were 5 ground truth higher order interactions then if we had 8 CorEx factors we would only look at the first 5. If we only had 3 CorEx factors then we would look at the first 3 and report the average score of the 3. Second row: Same as the first row except for the non-disjoint data. Third row: The average area under the precision-recall curve between each of the at-most k interactions found using CorEx with the closest ground truth interaction averaged across replicates and the number of latent factors for the disjoint data. Fourth row: Same as the third row except for the non-disjoint data. This largely confirms what we see in Figure \ref{fig:group_scores} but also confirms that each of the top $k$ latent factors is high quality indicating that we have good precision in identifying HOIs.}
    \label{fig:lf_scores}
\end{figure}

Another interesting takeaway from Figure \ref{fig:group_scores} is that when the higher order interactions do not differ that much between the two clusters, then the benefit of having the local partitioning decreases substantially. Despite this there still is a moderate benefit and a significant benefit when the interactions do not line up. Figure \ref{fig:lf_scores} is similar to Figure \ref{fig:group_scores} but instead focuses on the overall quality of the latent factors produced rather than looking at the coverage of the factors. I.e. Figure \ref{fig:lf_scores} represents the precision of the interactions found while Figure \ref{fig:group_scores} represents the recall of the interactions found.

\begin{figure}
    \centering
    \includegraphics[width=1.0\textwidth]{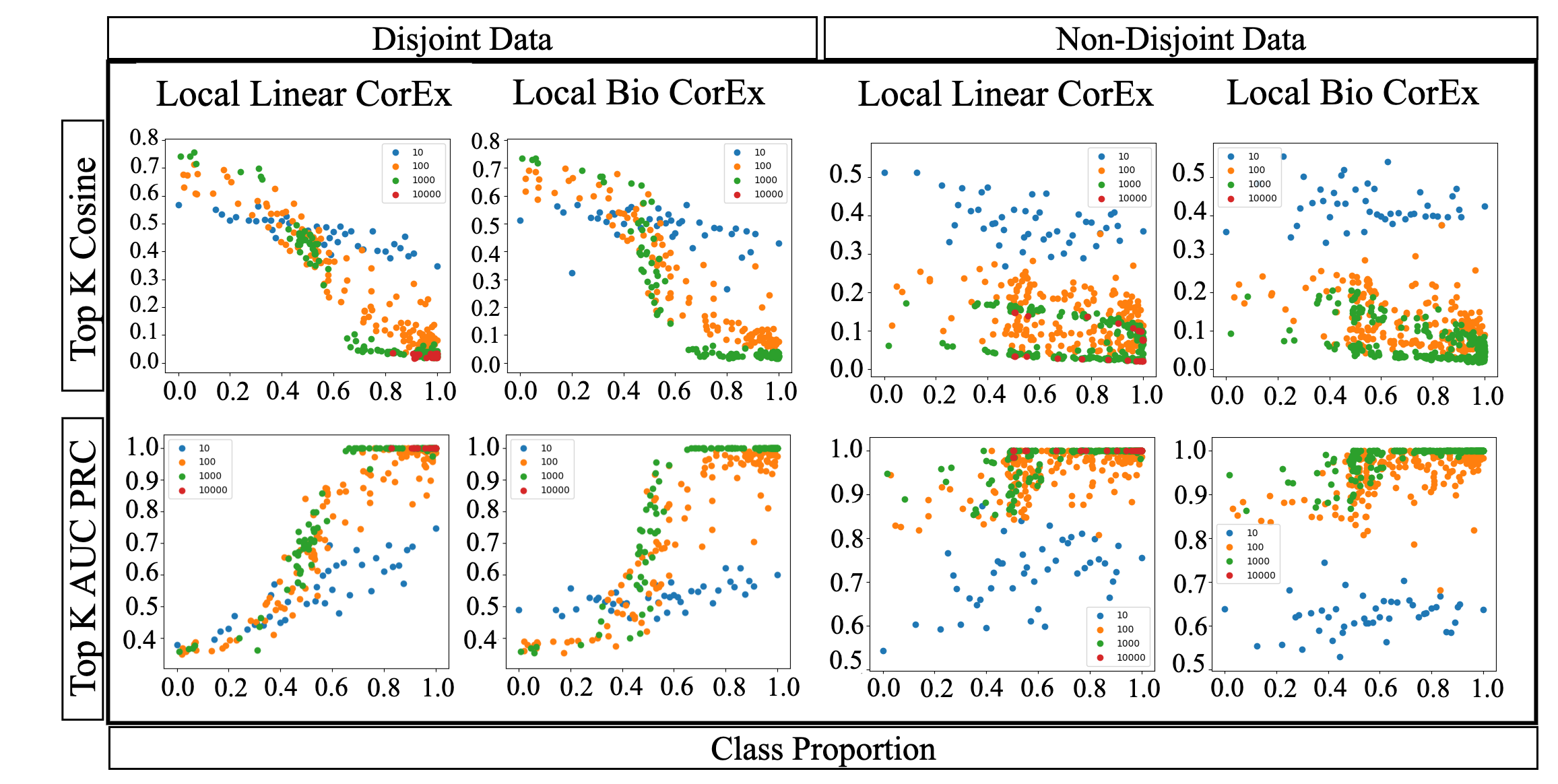}
    \caption{Similar to Figures \ref{fig:group_scores} and \ref{fig:lf_scores} except now the x-axis is the proportion of the class that it is being compared against. This shows that once a group has a little over \%60 purity of one sample type, there is a substantial benefit to using Local CorEx. This emphasizes that Local CorEx does not need perfectly pure partitions to work well.}
    \label{fig:class_prop_scores}
\end{figure}

Additionally, to better understand the relationship between the purity of a partition and the scores, we present Figure \ref{fig:class_prop_scores} where the x-axis now shows the proportion of the class that matches the group it is being compared against for ground truth HOIs. We see from the disjoint data that there is a substantial boost in performance when the class reaches about 60\% majority with the group it is being compared against. We also see from the non-disjoint data that it still does improve with class purity, but the improvement is much more modest.

\begin{figure}
    \centering
    \includegraphics[width=0.48\textwidth]{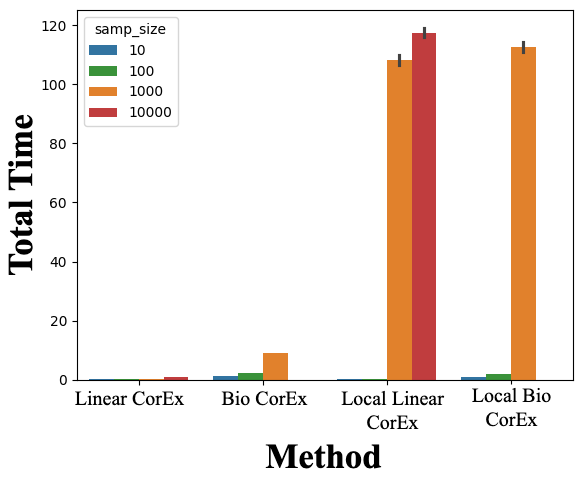}
    \includegraphics[width=0.48\textwidth]{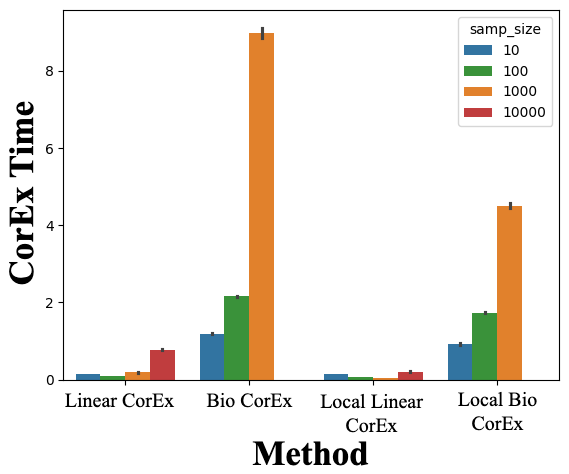}
    \caption{Computational comparison of Local CorEx to Bio and Linear CorEx. The average time in seconds to complete the 16 simulations is calculated for each setup. On the left side, we include the partitioning time, and on the right, we exclude the clustering time. Note that the majority of the computation time needed to run Local Linear CorEx is spent on clustering. In this work, we have chosen the clustering scheme as a set hyperparameter and have chosen not to explore other clustering schemes and leave this for further research. Of note, the clustering scheme more or less maxes out its time due to the landmark approach that PHATE uses after a sufficient number of samples are present. This is why despite the timing increases dramatically from a sample of size 100 to 1000 we don't see a similar increase from 1000 to 10000 (which is when the landmark approach kicks in by default).}
    \label{fig:timing}
\end{figure}

Finally, in Figure \ref{fig:timing} we see that Linear CorEx is faster than Bio CorEx. This holds true in the global and the local cases. We also learn that the majority of the time it takes to run Local CorEx is spent on partitioning the data. Due to the landmarking approach implemented in PHATE (that automatically is used once the number of samples is sufficient) the time it takes to cluster becomes more or less constant. This allows Local CorEx to scale well to large datasets. 

%\section{Communities Dataset} \label{appendix_communities_dataset}

    \begin{figure}
        \centering
        \includegraphics[width=.85\textwidth]{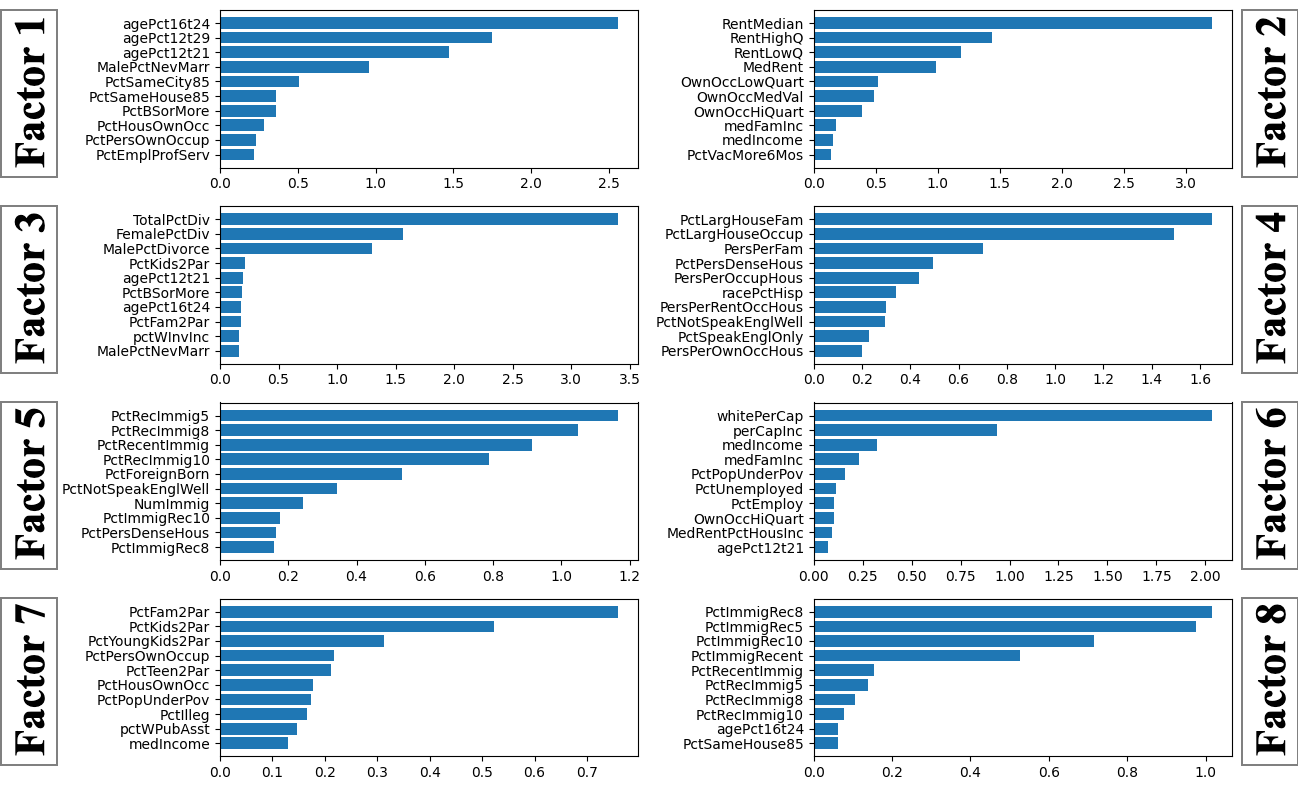}
        \caption{The first 8 Local CorEx Cluster 0 latent factors from the communities and crime dataset~\cite{misc_communities_and_crime_183} showing the mutual information between the factor and the original features sorted in descending order. This cluster consists of communities that on average are rural, on the lower end for median income, and hosts an older population. See \url{https://archive.ics.uci.edu/dataset/183/communities+and+crime} for descriptions of variable names.}
        \label{fig:group_0_factors}
    \end{figure}
    
    \begin{figure}
        \centering
        \includegraphics[width=.85\textwidth]{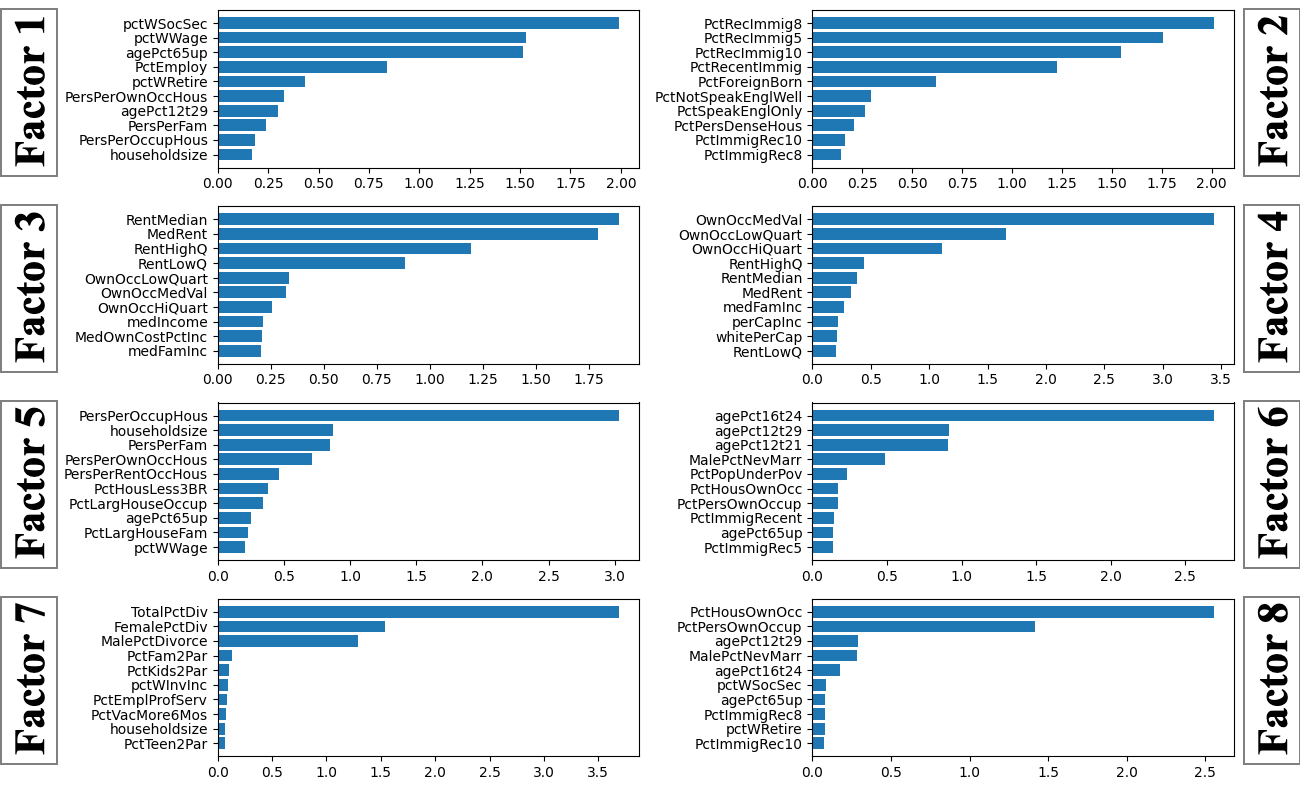}
        \caption{The first 8 Local CorEx Cluster 8 latent factors from the communities and crime dataset~\cite{misc_communities_and_crime_183} showing the mutual information between the factor and the original features sorted in descending order. This cluster consists of communities that on average are urban, on the lower to middle end for median income, and is not consistent for age above 65.  See \url{https://archive.ics.uci.edu/dataset/183/communities+and+crime} for descriptions of variable names.}
        \label{fig:group_8_factors}
    \end{figure}

\section{MNIST Dataset} \label{appendix_mnist_dataset}

Here we apply the same procedure used in Section \ref{sub:MNIST} to other clusters of the MNIST data.

\subsection{Cluster 15} \label{appendix_mnist_partition15}

\begin{figure}
    \centering
    \includegraphics[width=.9\textwidth]{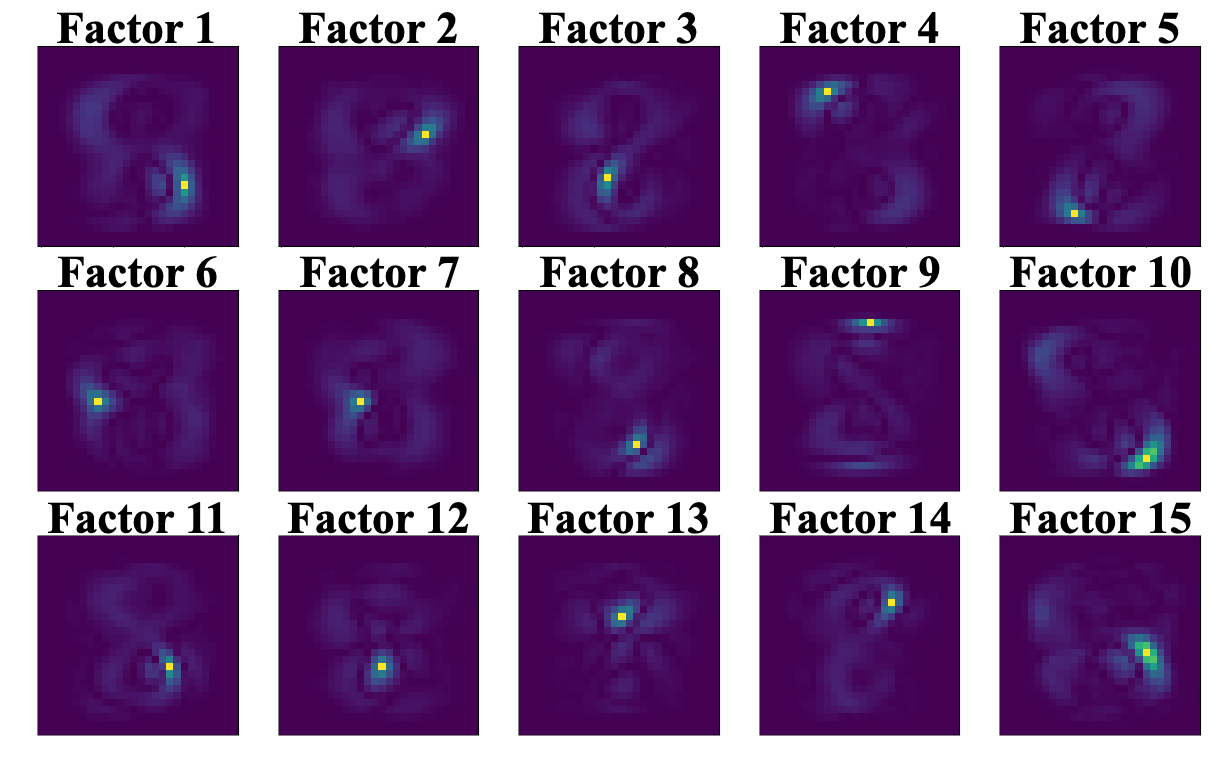}
    \caption{Visualizing the square-rooted mutual information between the first 15 Local CorEx factors trained on cluster 15 with the original features of the MNIST dataset.}
    \label{fig:mnist_group_15}
\end{figure}

 We repeat our analysis in Section~\ref{sub:MNIST} to cluster 15. Cluster 15 consists of 2011 8s, 681 3s, 125 5s, 103 9s, 40 2s, 14 0s, and less than ten of each of the other digits. We chose this cluster since it is fairly similar to cluster 16 in the sense that it is composed mainly of two digits (eights and threes). When examining Figure \ref{fig:mnist_group_15}, the first 2 factors seem to be related to the right-hand side curves of the digits, while the following three factors relate to the pixels capturing the left-hand side curves of the digits. It is clearly visible for each factor that the activated pixels relate to samples in the cluster.

\subsection{Cluster 23} \label{mnist_partition23}

\begin{figure}
    \centering
    \includegraphics[width=.9\textwidth]{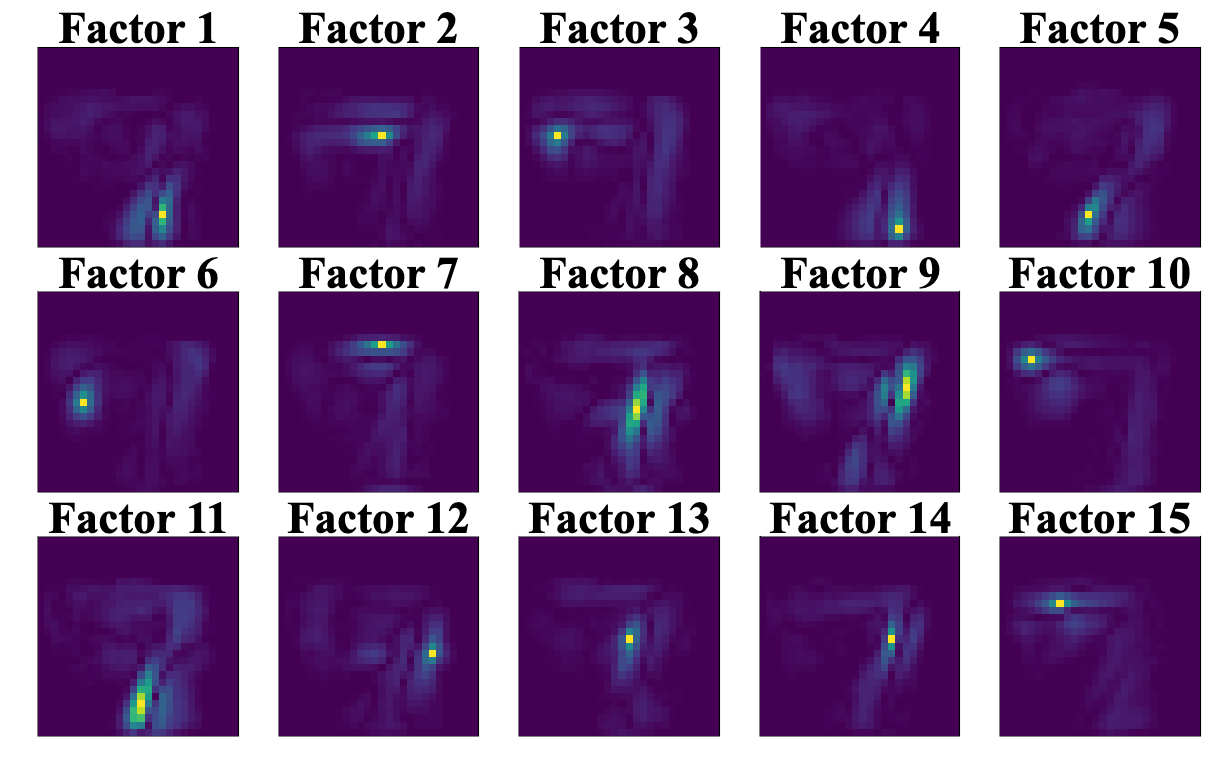}
    \caption{Visualizing the square-rooted mutual information between the first 15 Local CorEx factors trained on cluster 23 with the original features of the MNIST dataset.}
    \label{fig:mnist_group_23}
\end{figure}

The previous two clusters have consisted mostly of two classes. Cluster 23 contains almost exclusively 7s, and we include it to show that the method also works when partitions aren't pure in label despite being similar in input space. Cluster 23 consists of 1297 7s, 16 9s, and less than ten of each of the other digits. Notice in Figure \ref{fig:mnist_group_23} that practically all factors relate to either the stem of the seven or the top portion. 

\subsection{Global Analysis} \label{mnist_global}

\begin{figure}
    \centering
    \includegraphics[width=.9\textwidth]{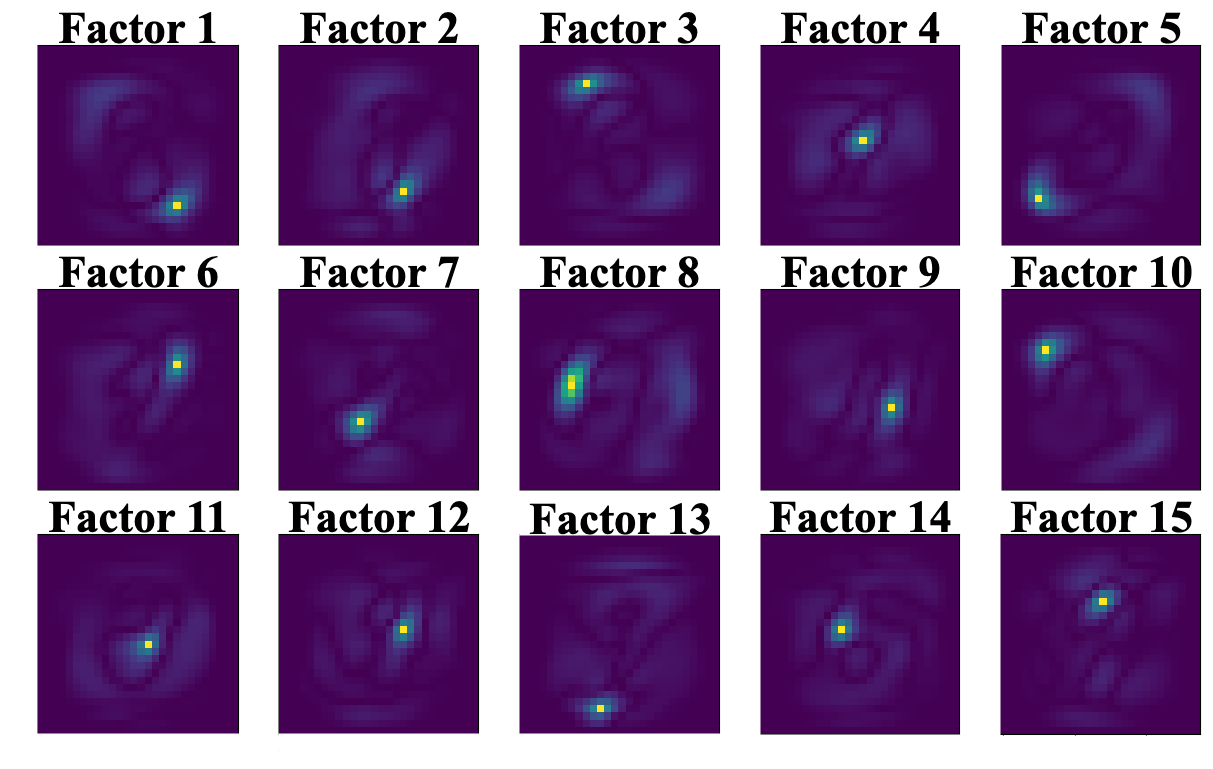}
    \caption{Visualizing the square-rooted mutual information between the first 15 Linear CorEx factors trained on the full dataset with the original features.}
    \label{fig:mnist_global}
\end{figure}

We now show how Linear CorEx performs when trained on the full dataset. The top 15 Local CorEx factors are presented in Figure \ref{fig:mnist_global}. The factors learned are less easy to fully understand since the context they reside in is much larger than the contexts for the local clusters in the previous sections. Despite this, we see that it groups sets of pixels together that seem reasonable and even later factors like factors 13 and 14 could be associated with 7s and 5s respectively. 

\section{Neural Network} \label{neural_network}

\begin{figure*}
            \centering
            \includegraphics[width=.48\linewidth]{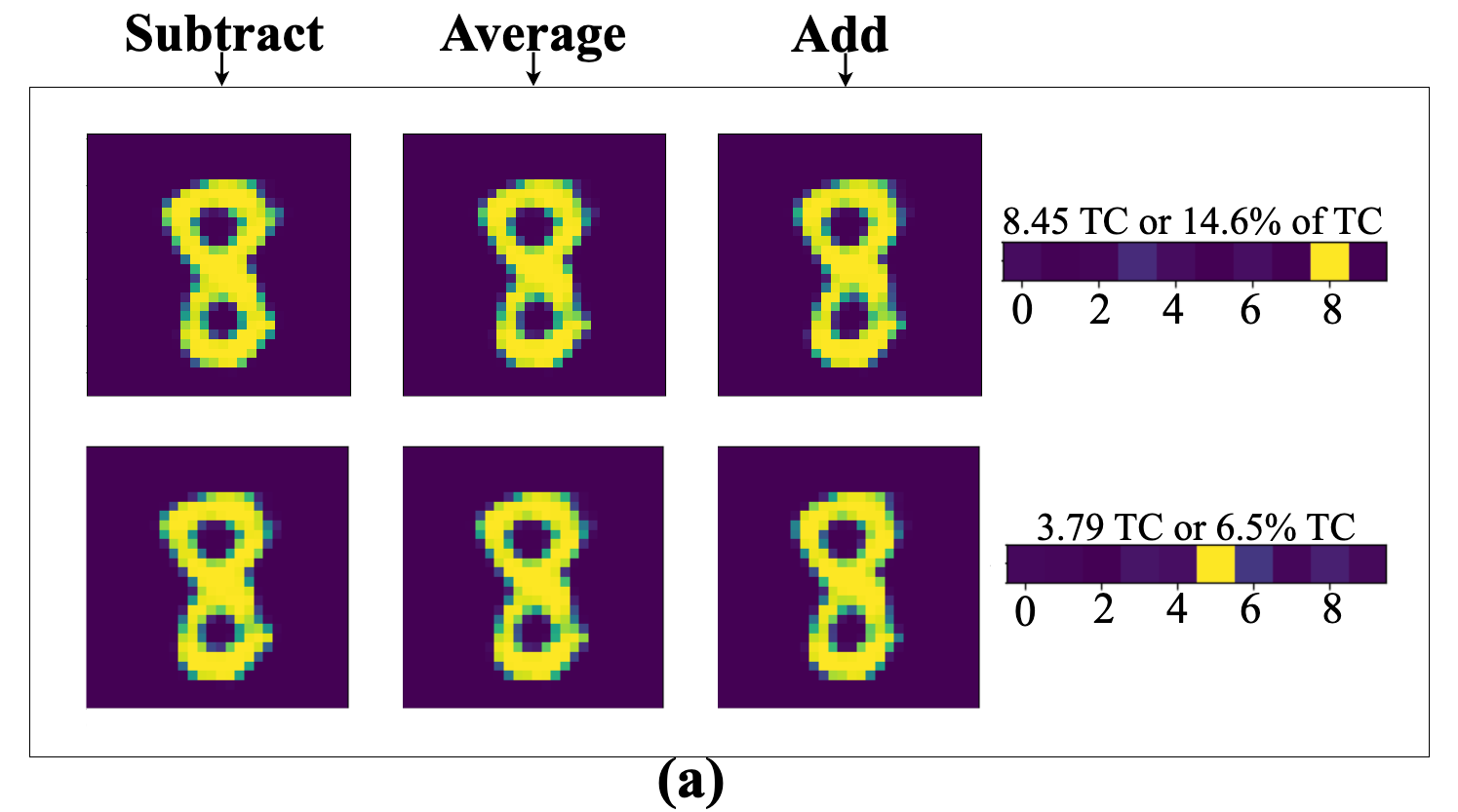}
            \includegraphics[width=.48\linewidth]{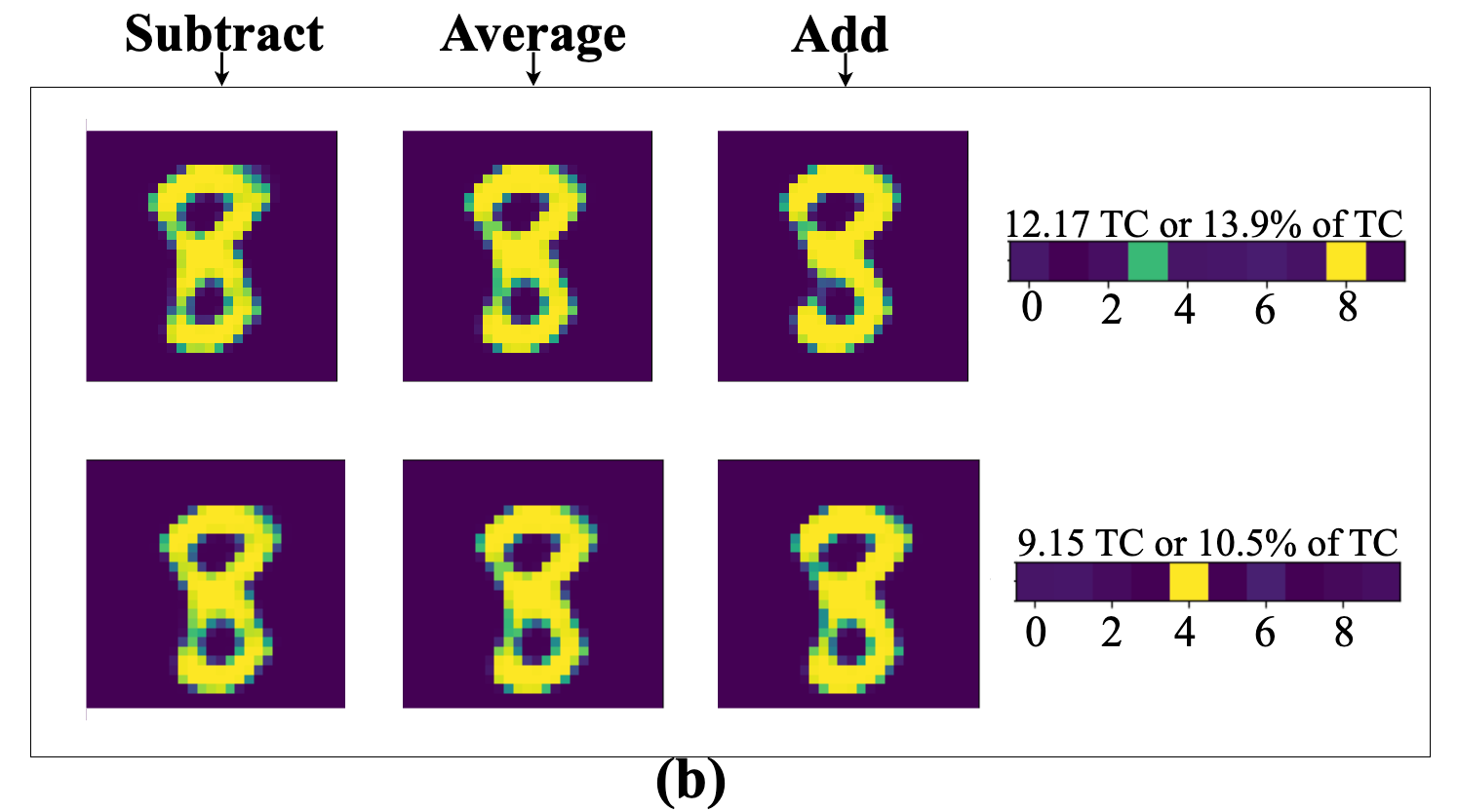}
            \caption{Visualizing the effect of perturbing the average hidden state representations of cluster 12. \textit{(a)} The plots are associated with perturbing the H1 representation. This first row is perturbing the first CorEx factor and the second row is associated with the second CorEx factor for the respective hidden representation. \textit{(b)} The plots here are associated with perturbing the H2 representation. The first row is perturbing the first CorEx factor and the second row is associated with the second CorEx factor for the respective hidden representation. For each group of plots, the leftmost column image is generated by subtracting the mutual information between the CorEx factor and the hidden nodes from the average representation. The second column image gives the average hidden state representation for partition 16. The third column image is generated by adding the mutual information between the CorEx factor and the hidden nodes from the average representation. Finally, the rightmost column plots the mutual information between the CorEx factor and the model logits. This analysis gives us a visual intuition for what role the grouped hidden nodes found using Local CorEx play.}
            \label{fig:reconstructed_hs_group_12}
        \end{figure*}

For completeness, we include analysis on an additional cluster. We visualize the effects of groups of hidden nodes in the same manner as done in Section \ref{subsub:vis_hidden_node}. Following it we generate Figure \ref{fig:reconstructed_hs_group_12}. From this figure we see that perturbing the H1 representation according to the mutual information with the first Local CorEx factor affects the intensity of the pixels in the right side of the bottom curve of the eight. Perturbing the H1 representation according to the mutual information with second Local CorEx factor affects the thickness of the pixels where the two circles in the eight meet. Perturbing the H2 representation according the first Local CorEx factor effects the pixels on the bottom left side of the eight. Perturbing the second Local CorEx factor effects the pixels on left hand side of the eight.

\begin{figure}
            \centering
            \includegraphics[width=.9\textwidth]{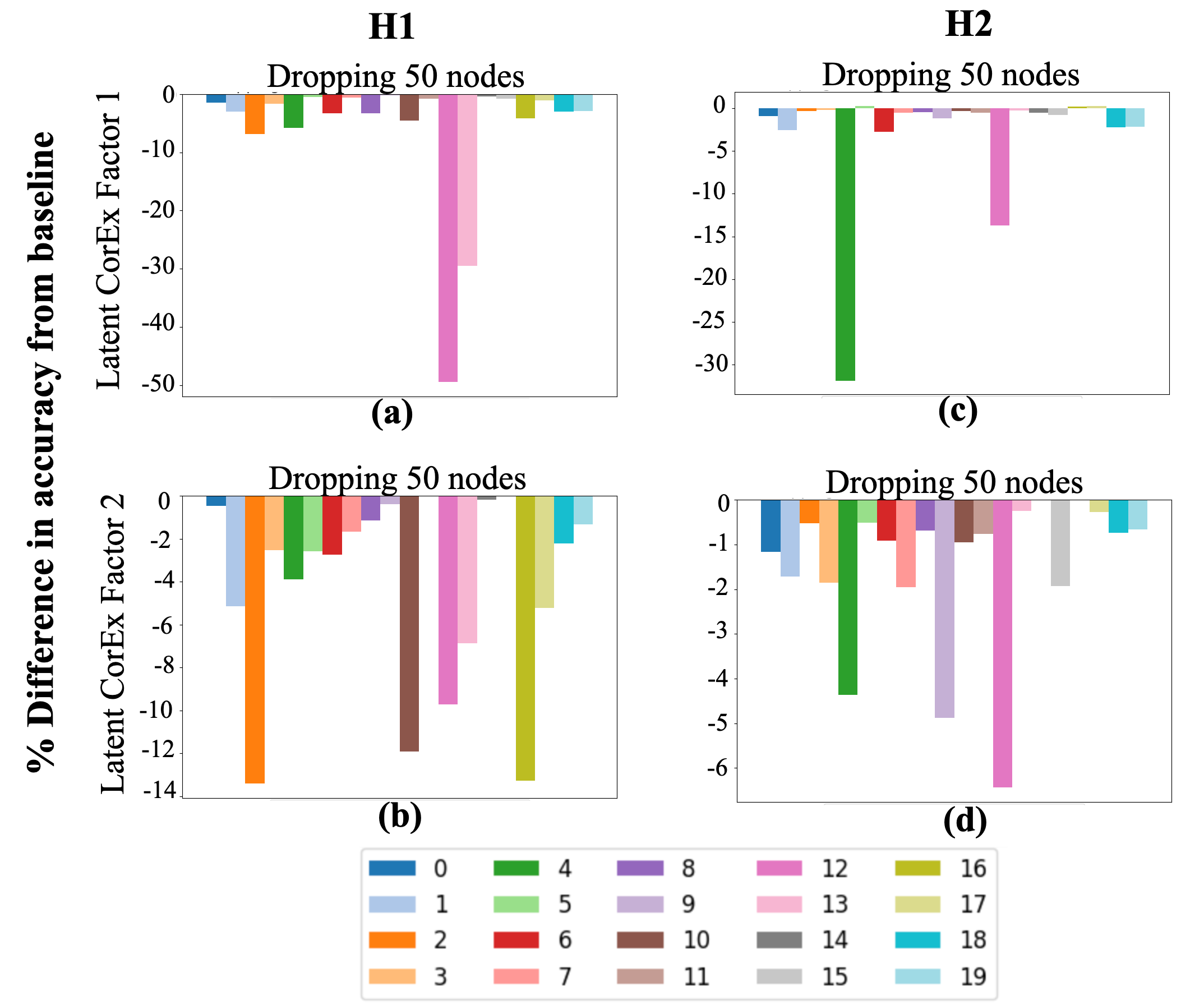}
            \caption{\textit{(a)} Difference in classification accuracy between the unaltered model and the model after deleting 50 hidden nodes in the first hidden layer with the highest mutual information with the first CorEx factor associated with cluster 12. This factor is associated with the logit for classifying 8s and the nodes associated with this factor are critical for accurately classifying 8s and 3s since those are the dominant classes for clusters 12 and 13. \textit(b) Same as \textit{(a)} except when using the second CorEx factor to determine the 50 hidden nodes to delete. This factor is associated with the logit responsible for classifying 5s and has a mixed effect. \textit(c) Difference in classification accuracy between the unaltered model and the model after deleting 50 hidden nodes in the second hidden layer with the highest mutual information with the first CorEx factor. This factor is associated with the logit for classifying 8s and is critical for classifying clusters 4 and 12. \textit(d) Difference in classification accuracy between the unaltered model and the model after deleting 50 hidden nodes in the second hidden layer with the highest mutual information with the second CorEx factor. This factor is associated with the logit for classifying 4s and affects clusters 4, 9, and 12 the most. Note that the y-axis scale differs for each plot. The first CorEx factors for both H1 and H2 are associated with the logit for classifying eights. The groups of nodes associated with these factors are critical for accurately classifying the digits in clusters 2, 10, and 16 which contain mainly 4s and 9s. The second CorEx factors for both H1 and H2 correlate with logits for digits 0, 2, and 6 and seem to encode information less necessary for accurately classifying one particular group.}
            \label{fig:dropping_hidden_nodes_group_12}
        \end{figure}

    When we delete the hidden nodes in the same manner as done in Section \ref{subsub:quantify_node_effect} we generate the plots shown in Figure \ref{fig:dropping_hidden_nodes_group_12}. In it, we see that the nodes associated with the first factor for the H1 representation almost exclusively affect clusters 12 and 13, which have majority classes of eights. This makes sense since the Local CorEx factor is associated with the logit responsible for classifying 8s. When we delete the second factor for the H1 representation we see that it affects several clusters, but the ones most affected are clusters 2, 10, 12, 13, and 16. All of these clusters consist of fours, nines, eights, and threes and in Figure \ref{fig:reconstructed_hs_group_12} we see that this factor is associated with the thickness where the two loops of the eight meet (in particular, it adds pixels to the bottom right portion). Removing the nodes responsible for this information could make fours and nines look more like eights and eights more like nines.

    For the nodes associated with the first factor for the H2 representation, we see that clusters 4 and 12 are the most affected. When we look at Figure \ref{fig:reconstructed_hs_group_12} to see the effect of this factor we see it primarily affects the lower loop's pixels on the left side. These pixels line up for clusters 4 and 12. If we look at cluster 13, which is the only other cluster where the majority is either a three or an eight we see that the bottom loop is more to the left than that of cluster 12 explaining why this cluster isn't also affected. When we look at the second factor we that the clusters most strongly affected are 4, 9, and 12. When we consult Figure \ref{fig:reconstructed_hs_group_12} we see that this factor affects both the top and bottom loops on the left-hand side. It thus isn't surprising that groups 4 and 12 are affected since they contain many threes, but when we look at cluster 9 we see that those pixels could make a seven look more like a nine. 

\begin{figure*}
    \centering
    \includegraphics[width=.9\linewidth]{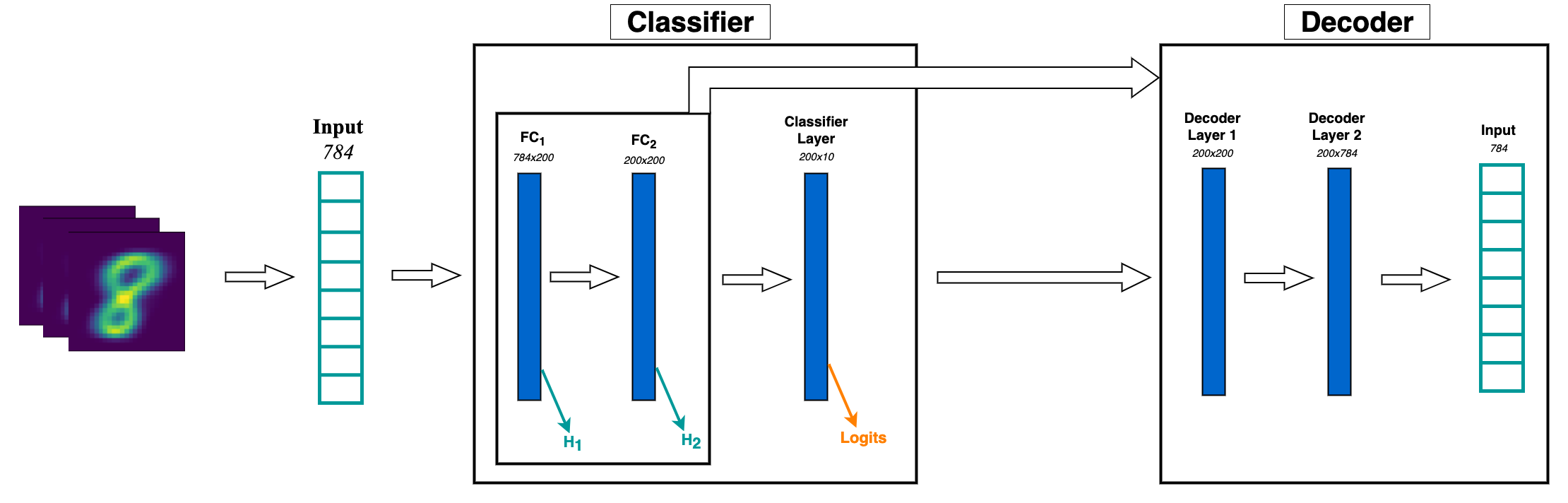}
    \caption{Neural Network architecture. MNIST digit samples are first flattened into vectors of size 784 before being passed into the neural network. As the inputs are passed through each layer in the classifier we save each internal representation. After a classifier is fully trained we also train a decoder model that uses the first layers of the classifier as an encoder whose weights are frozen during the decoder's training.}
    \label{fig:model_architectures}
\end{figure*}

\begin{figure*}
    \centering
    \includegraphics[width=.9\linewidth]{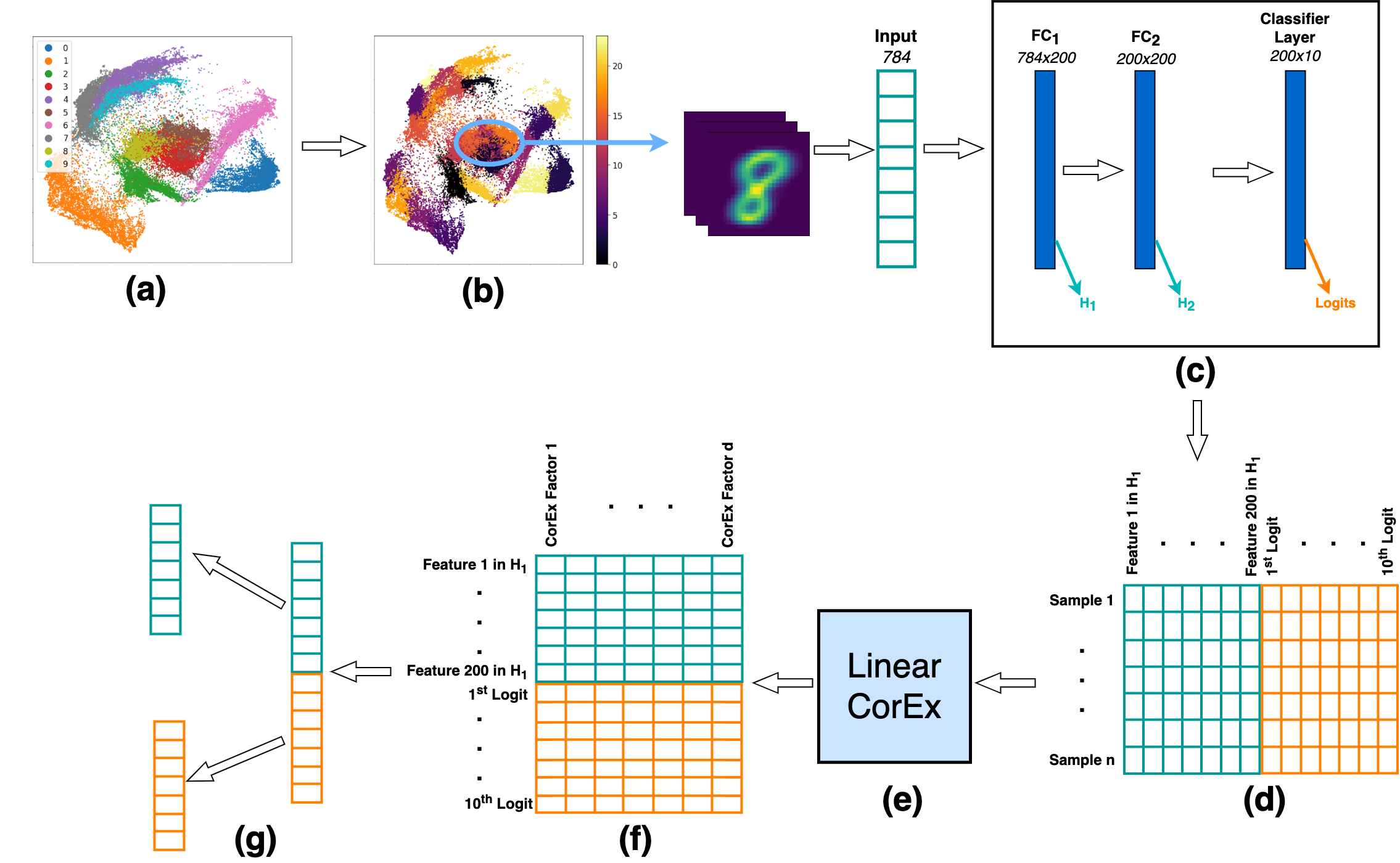}
    \caption{Running Local CorEx on a hidden representation of the inputs concatenated with their respective logits. \textit{(a)} PHATE embedding of the MNIST data, colored by class. \textit{(b)}  $k$-means clustering is applied to the PHATE embedding to generate the partition. \textit{(c)} MNIST digit samples are flattened into vectors of size $784$ before being passed into the neural network. As the inputs are passed through each layer in the classifier we save each internal representation. \textit{(d-e)} The hidden layer outputs (e.g. $H_1$ hidden state) are concatenated with their respective logits and passed through Linear CorEx. We repeat this for the hidden state $H_2$ with their respective logits. \textit{(f)} We visualize the mutual information between the learned CorEx factors and the concatenated features with their respective logit. \textit{(g)} We pick one latent factor to continue to explore and split back into the concatenated feature vector and its respective logits.
    This allows us to transform hidden state representations into reconstructed inputs and identify groups of hidden nodes to delete.}
    \label{fig:neural_net_processing}
\end{figure*}

\begin{figure}
    \centering
    \includegraphics[width=1\linewidth]{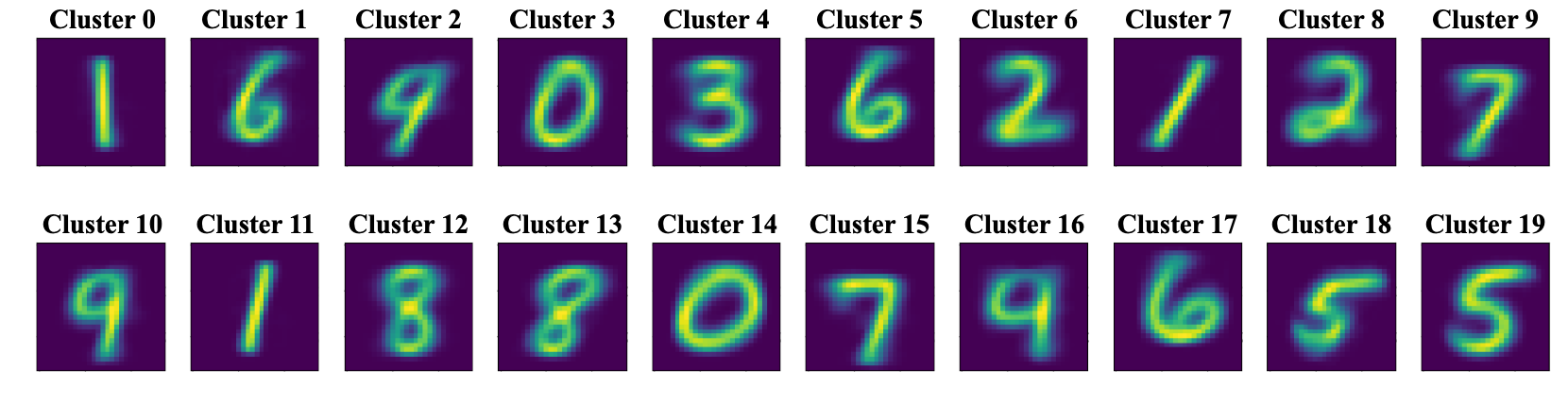}
    \caption{Visualization of the average samples in each of the 20 partitions for the MNIST test dataset used in Section \ref{sub:neural_net}.}
    \label{fig:mnist_test_partitions}
\end{figure}

\begin{table}[t]
    \caption{Baseline classifier accuracies across clusters on the test split of the MNIST dataset. With the accuracies, for each cluster, we include the number of each label for the 3 most represented digits in the cluster. From this we can see that the baseline model performs well across clusters.}
    \label{tab:mnist_classifer_performance}
    \begin{center}
    \begin{small}
    \begin{sc}
    \begin{tabular}{c|c|l}
    \toprule
        Clusters & Accuracy & Cluster Composition - Top 3\\
        \hline
        0 & 98.60 & 1: 404,  4: 9,    7: 8\\
        1 & 96.57 & 6: 200,  5: 11,   2: 6\\
        2 & 98.08 & 4: 314,  9: 249,  8: 5\\
        3 & 98.15 & 0: 406,  2: 11,   5: 6\\
        4 & 98.20 & 3: 799,  5: 172,  8: 78\\
        5 & 99.23 & 6: 384,  4: 3,    0: 1\\
        6 & 97.80 & 2: 527,  7: 7,    1: 5\\
        7 & 97.49 & 1: 338,  7: 9,    2: 5\\
        8 & 99.54 & 2: 433,  9: 1\\
        9 & 97.07 & 7: 461,  9: 22,   2: 20\\
        10 & 97.86 & 9: 475,  4: 324,  7: 29\\
        11 & 99.49 & 1: 378,  7: 7,    4: 4\\
        12 & 98.00 & 8: 468,  3: 177,  5: 14\\
        13 & 99.51 & 8: 379,  9: 12,   3: 8\\
        14 & 99.64 & 0: 557,  2: 2,    6: 2\\
        15 & 98.26 & 7: 496,  9: 7,    3: 6\\
        16 & 98.88 & 4: 317,  9: 212,  7: 4\\
        17 & 99.17 & 6: 359,  4: 2,    9: 1\\
        18 & 97.78 & 5: 247,  8: 11,   0: 4\\
        19 & 98.90 & 5: 436,  3: 8,    8: 6\\
        \bottomrule
    \end{tabular}
    \end{sc}
    \end{small}
    \end{center}
\end{table}

\section{Local CorEx in practice}
\label{sec:hyperparameters}
Below we provide some general guidance for tuning hyperparameters for running Local CorEx. Hyperparameters include: the number of dimensions to include in the PHATE embedding of the data for clustering, the number of clusters for $k$-means, the number of CorEx latent factors to learn, and for each CorEx factor the threshold for determining when a feature is part of a predicted HOI. 

To find the number of dimensions of the PHATE embedding for clustering, we have found that in most cases, unless there are few observed features, that a number around 10 works well. If desired, a principled way to determine the number of dimensions would be to check the loss of the final MDS stage as a function of embedding dimension. To determine $k$ for $k$-means we recommend constructing a 2D PHATE plot of the data to visually inspect it to determine a baseline number of clusters needed. As shown in Appendix \ref{appendix_ablation_study} Local CorEx doesn't suffer greatly from smaller sample sizes so long as there are 50 to 100 samples in the cluster. For determining the number of CorEx factors to be learned, you can examine the total correlation explained by each factor and only keep factors where the contribution isn't nominal. For thresholding learned HOIs, we discovered that visualizing them where you can see the mutual information context is helpful in determining where to stop. One could construct a scree plot and determine a threshold in that manner if desired.

%%%%%%%%%%%%%%%%%%%%%%%%%%%%%%%%%%%%%%%%%%%%%%%%%%%%%%%%%%%%%%%%%%%%%%%%%%%%%%%
%%%%%%%%%%%%%%%%%%%%%%%%%%%%%%%%%%%%%%%%%%%%%%%%%%%%%%%%%%%%%%%%%%%%%%%%%%%%%%%

\end{document}